\documentclass[letterpaper]{article} 
\usepackage{aaai25}  
\usepackage{times}  
\usepackage{helvet}  
\usepackage{courier}  
\usepackage[hyphens]{url}  
\usepackage{graphicx} 
\usepackage{natbib}  
\usepackage{caption} 
\usepackage{algorithm}
\usepackage{algorithmic}

\usepackage{newfloat}
\usepackage{listings}
\DeclareCaptionStyle{ruled}{labelfont=normalfont,labelsep=colon,strut=off} 
\floatstyle{ruled}
\newfloat{listing}{tb}{lst}{}
\floatname{listing}{Listing}
\usepackage{booktabs}

\title{Machine Learning and Public Health:\\
Identifying and Mitigating Algorithmic Bias through a Systematic Review\thanks{Extended version of the paper accepted at the AAAI/ACM Conference on AI, Ethics, and Society (AIES 2025), including an appendix. Licensed under Creative Commons Attribution (CC BY 4.0).}}

\author {
    Sara Altamirano,
    Arjan Vreeken,
    Sennay Ghebreab
}
\affiliations {
    Informatics Institute, University of Amsterdam, The Netherlands\\
    \{s.e.altamirano, a.vreeken, s.ghebreab\}@uva.nl
}
\nocopyright
\begin{document}

\maketitle

\begin{abstract}
Machine learning (ML) promises to revolutionize public health through improved surveillance, risk stratification, and resource allocation. However, without systematic attention to algorithmic bias, ML may inadvertently reinforce existing health disparities. We present a systematic literature review of algorithmic bias identification, discussion, and reporting in Dutch public health ML research from 2021 to 2025. To this end, we developed the Risk of Algorithmic Bias Assessment Tool (RABAT) by integrating elements from established frameworks (Cochrane Risk of Bias, PROBAST, Microsoft Responsible AI checklist) and applied it to 35 peer-reviewed studies. Our analysis reveals pervasive gaps: although data sampling and missing data practices are well documented, most studies omit explicit fairness framing, subgroup analyses, and transparent discussion of potential harms. In response, we introduce a four-stage fairness-oriented framework called ACAR (Awareness, Conceptualization, Application, Reporting), with guiding questions derived from our systematic literature review to help researchers address fairness across the ML lifecycle. We conclude with actionable recommendations for public health ML practitioners to consistently consider algorithmic bias and foster transparency, ensuring that algorithmic innovations advance health equity rather than undermine it.
\end{abstract}

%

\section{Introduction}

Machine Learning (ML) is transforming Public Health (PH) through accurate prediction, real-time monitoring, and data-driven decision-making \citep{mhasawade2021machine, rajkomar2018ensuring, wiemken2020machine, benke2018artificial, jiang2017artificial}. As PH systems adopt ML, these technologies become essential. Yet ethical governance lags behind. Despite progress in ML fairness, its application in PH—especially for vulnerable populations—remains limited. Without intervention, algorithmic bias (AB) can exacerbate health disparities, misallocate resources, and reinforce care barriers \citep{fletcher2021addressing, flores2024addressing, gianfrancesco2018potential}. 
These risks place PH+ML research within the broader domain of Responsible Artificial Intelligence (AI). The challenge is sociotechnical: integrating algorithmic decision-making (ADM) into real-world settings where equity, transparency, and accountability are critical. In high-stakes fields like PH, where decisions affect entire populations, unreported or misunderstood bias carries significant risks. As ML adoption grows, it becomes urgent to assess how AB is embedded in PH+ML research and whether it is adequately addressed by the research team \citep{xu2022algorithmic, thomasian2021advancing}. 

This study systematically investigates how AB is identified, discussed, and reported in Dutch PH+ML research. While ML increasingly informs PH, it remains unclear whether researchers recognize and communicate AB risks. We highlight reporting gaps and advocate for transparent, standardized, fairness-aware practices in a domain with direct population-level impact.

To this end, we conduct a systematic literature review (SLR) of Dutch PH+ML studies. We extract metadata, including ML tasks, datasets, and performance metrics. Using our Risk of Algorithmic Bias Assessment Tool (RABAT), we code how AB risks, subgroup vulnerabilities, and fairness considerations are addressed—or not. Alongside this retrospective assessment, we introduce the ACAR framework (Awareness, Conceptualization, Application, Reporting): a forward-looking guide to help PH+ML researchers assess and address AB across the research lifecycle. ACAR translates SLR insights into guiding questions that embed fairness and accountability from conception to reporting, enabling ethical, context-sensitive system design.

The Netherlands is a highly relevant setting, with strong PH and ML innovation supported by advanced health infrastructure and digitalization. Despite universal healthcare, disparities persist—especially among ethnic minorities and socioeconomically disadvantaged groups, who face care barriers and higher burdens of chronic and psychiatric conditions \citep{ikram2014disease, ilozumba2022ethnic, kroneman2016netherlands}. While focused on Dutch PH+ML, this study offers globally relevant insights for countries seeking to ensure ML promotes, rather than undermines, health fairness and ethical accountability.

In summary, this study advances Responsible AI by (1) conducting an SLR of AB reporting in Dutch PH+ML research, (2) developing and applying RABAT to identify key gaps, and (3) introducing ACAR as a practical fairness framework. These contributions show that addressing ethical concerns in PH+ML requires more than technical fixes; it demands rethinking how we evaluate, design, and report algorithms for the public good.

\section{Background}

\subsection{Key Definitions}

ADM refers to computational systems, often ML-powered, that support or automate decisions with real-world consequences \citep{burrell2016how, mittelstadt2016ethics}. These systems are increasingly used in high-stakes domains such as credit scoring \citep{altman1998credit}, stock market prediction \citep{patel2015predicting}, educational risk alerts \citep{arnold2012course}, recidivism prediction \citep{angwin2016machine}, digital public services \citep{margetts2017government}, and PH surveillance \citep{Ginsberg_2009, hillebrand2020mobisenseus}. While enabling large-scale pattern detection and efficiency gains, ADM also risks amplifying structural disparities. For instance, models may rely on proxy variables like ZIP codes that correlate with race, producing biased outcomes, and their opacity can obscure accountability \citep{Barocas_2016, burrell2016how}. These risks are well documented, particularly in public sector ADM, where opaque systems have exacerbated existing inequalities \citep{eubanks2018automating}.

One of the main risks of ADM systems is AB: systematic patterns in data, model design, or deployment that yield unfair or discriminatory outcomes, often disadvantaging particular individuals or groups. It arises from sources such as sampling gaps, flawed feature selection, and inherited historical inequities \citep{mehrabi2021survey, suresh2019framework}. Documented cases include racial bias in facial recognition systems \citep{buolamwini2018gender}, recidivism risk scores \citep{angwin2016machine}, clinical prediction tools based on healthcare costs \citep{obermeyer2019dissecting}, and stereotyping in GenAI image generation \citep{ferrara2024genai}. AB may be passive—inherited from data—or active, introduced during design. Detecting it can reveal structural inequities and support mitigation. Risks include misallocation of resources, flawed predictions, and harm to marginalized groups. AB reflects broader structural injustices and calls for responses that combine technical mitigation with societal interventions.

In contrast, algorithmic fairness (AF) refers to the principles, definitions, and interventions aimed at preventing or mitigating the harms of AB. AF encompasses both individual and group level objectives grounded in social and ethical judgments about justice and discrimination \citep{barocas-hardt-narayanan, mehrabi2021survey}. AF interventions range from reactive audits to proactive fairness-aware optimization, and their application must be context-sensitive to align with domain-specific values \citep{caton2024fairness, binns2018fairness, selbst2019fairness}. Formal AF definitions include individual fairness \citep{dwork2012fairness}, statistical parity \citep{feldman2015certifying}, equalized odds \citep{hardt2016equality}, predictive parity \citep{chouldechova2017fair}, and counterfactual fairness \citep{kusner2017counterfactual}. These metrics often involve trade-offs and require careful consideration within each application context \citep{verma2018fairness}. Moreover, effective AF can enhance transparency, reduce disparities, and foster public trust \citep{holstein2019improving}. However, over-reliance on a single metric risks obscuring deeper systemic inequities \citep{corbett-davies2023measure} or enabling “fairwashing”—superficial fairness claims without substantive change \citep{aivodji2019fairwashing}. Meaningful AF demands stakeholder engagement, institutional commitment, and ongoing evaluation, especially in PH, where fairness challenges are complex and multifaceted.

\subsection{Reporting Bias in PH+ML Research}
Guidelines for reporting AB in PH+ML remain limited. While \citet{thomasian2021advancing} outline pipeline bias strategies specific to PH data systems, most available frameworks are adapted from clinical or general ML contexts. \citet{vollmer2020machine} propose 20 critical questions via the TREE checklist for clinical ML, and \citet{rajkomar2018ensuring} offer a widely cited fairness checklist rooted in distributive justice principles. \citet{fletcher2021addressing} contribute three structured criteria—appropriateness, fairness, and bias—for evaluating ML systems in global health.

Documentation tools such as Model Cards and Datasheets promote transparency but lack PH+ML-specific focus \citep{mitchell2019model, gebru2021datasheets}. Healthsheets \citep{rostamzadeh2022healthsheet} adapt datasheets to healthcare, using expert input to foreground dataset-level bias. BEAMRAD \citep{galanty2024assessing} evaluates transparency in medical imaging and signal datasets, linking poor documentation to downstream risks. However, these tools rarely address subgroup harms or equity challenges specific to PH contexts.

Several EQUATOR Network frameworks aim to improve reporting quality in biomedical research. TRIPOD+AI and PROBAST+AI offer detailed guidance on reporting and assessing risk of bias and applicability in clinical prediction models. However, they provide only high-level recommendations regarding subgroup analyses and do not operationalize fairness evaluations or subgroup-specific metrics \citep{collins2024tripodai, moons2025probastai}. STARD-AI, CLAIM, CONSORT-AI, and SPIRIT-AI target diagnostic or interventional studies but address AB in population-level ML \citep{ibrahim2021stardai, mongan2020claim, liu2020consortai}. Other EQUATOR tools—PRISMA \citep{page2021prisma}, SPIRIT \citep{chan2013spirit}, STROBE \citep{vonelm2007strobe}, CARE \citep{gagnier2013care}, COREQ \citep{tong2007coreq}, SRQR \citep{obrien2014standards}, STARD \citep{Bossuyt_2015}, TRIPOD \citep{collins2015tripod}, CHEERS \citep{husereau2013cheers}, and ARRIVE \citep{kilkenny2010arrive}—support transparency but overlook fairness considerations, subgroup calibration, or bias mitigation.

Beyond clinical guidelines, frameworks like FUTURE-AI \citep{lekadir2025futureai}, IEEE 7003 \citep{ieee2025bias}, and WHO/UNESCO ethics guidance \citep{who2021ethics} emphasize responsible AI development, but remain high-level and lack actionable tools for AB reporting in PH+ML research.

Open-source toolkits such as AIF360 \citep{bellamy2019aif360}, Fairlearn \citep{bird2020fairlearn}, Themis \citep{galhotra2017fairness}, fairmodels \citep{wisniewski2022fairmodels}, the What-If Tool \citep{wexler2019whatif}, FairTest \citep{tramer2017fairtest}, and Aequitas \citep{saleiro2018aequitas} provide general fairness metrics and mitigation techniques. While Aequitas includes a “fairness tree” adapted for population-level settings, none of these tools support PH-specific bias detection or subgroup-sensitive calibration.

In sum, existing reporting frameworks promote general transparency but lack mechanisms for identifying, measuring, or mitigating AB in PH+ML research.

\subsection{Dutch Context and Governance Landscape}

The Netherlands is a highly relevant setting for studying AB in PH+ML research, with advanced data systems, high Electronic Health Records (EHR) coverage, and a diverse population \citep{kroneman2016netherlands}. ML is increasingly used in Dutch public institutions, including PH, supporting ADM across policy domains. However, structural disparities persist. Ethnic minorities and disadvantaged groups face higher burdens of chronic disease and barriers to mental health care, while migrant populations report unmet cultural and linguistic needs \citep{ikram2014disease, ilozumba2022ethnic, teunissen2015mental}. Without explicit AB attention, ML risks amplifying these inequities. 

Recent governance reforms reflect growing awareness, regulatory action, and progress in technical tools for responsible AI. In 2022, Statistics Netherlands (CBS) replaced its binary “Western/non-Western” classification with more nuanced categories, reshaping how sensitive attributes are defined and operationalized \citep{cbs2022}. The Netherlands also maintains a national Algorithm Register, which documents deployed ADM systems in domains such as youth care, benefits, and PH surveillance \citep{algoritmeregister2024}. While ML increasingly informs PH decision-making, systematic AB reporting remains rare. One exception is \citet{holstege2025auditing}, who audited bias in an administrative risk profiling system. Although Dutch and EU governance frameworks emphasize transparency, they do not mandate AB reporting in PH+ML research. The Dutch AI Impact Assessment and EU AI Act prioritize high-risk domains such as law enforcement, employment, and healthcare, with limited attention to PH or AB reporting \citep{miniwm2024aiimpact, euai2024}. The Dutch Vision on Generative AI broadly mentions health and fundamental rights but does not address PH or AB directly  \citep{minbzk2024genai}.

\subsection{Positioning This Study}
A growing body of literature has explored AB and AF in PH. \citet{mhasawade2021machine} identified gaps in fairness-aware ML and emphasized the need to incorporate structural determinants alongside individual factors. \citet{delgado2022bias} reviewed COVID-19 AI systems, highlighting recurring biases in data representativeness, demographic omission, and validation practices. Other reviews propose PH-relevant AF frameworks: \citet{sikstrom2022conceptualising} conceptualize fairness as a multidimensional construct, while \citet{chin2023guiding} centers health and health care equity for patients and communities as the goal, specifically within the wider context of structural racism and discrimination. \citet{char2020identifying} offer a stage-based ethical reflection framework for healthcare ML, outlining questions across development, implementation, and oversight—though not tailored to PH concerns. \citet{morgenstern2020predicting} conducted a scoping review of 231 ML prediction studies in PH, noting transparency and calibration gaps but not assessing AB or subgroup fairness.

Empirical studies further show how design and deployment choices can reinforce inequities. \citet{obermeyer2019dissecting} found that using healthcare cost as a proxy for illness systematically underestimated patient needs. \citet{tsai2022algorithmic} and \citet{flores2024addressing} identified structural and data-related fairness gaps in PH forecasting and surveillance.

Unlike conceptual or secondary reviews, this study systematically examines how AB risks are identified, discussed, and reported in Dutch PH+ML research. Based on the findings, we present the ACAR framework to translate reporting gaps into actionable fairness-aware guidance for PH+ML.

\section{Methods}
\label{sec:methods}

\subsection{Study Design}
This study addresses the growing use of ML in PH research, the risks of AB to vulnerable populations, the limited integration of AF, and the Netherlands’ position as a scientifically advanced yet unequal health system. We conducted an SLR to analyze trends in AB reporting and AF consideration in Dutch PH+ML studies. Following interdisciplinary SLR guidelines \citep{carrera2022conduct}, the review was structured for rigor, replicability, and comprehensiveness. Our goal was to examine how AB risk is addressed across the research process, focusing on impacts on minorities and disadvantaged groups. Our main research question was: To what extent is AB risk identified, discussed, and reported in Dutch PH+ML research from 2021–2025?

The review targeted three categories—\textit{extent of bias discussion}, \textit{identification of subgroups at risk}, and \textit{reporting transparency}—assessed via RABAT. Scope was defined using the PICOC framework \citep{carrera2022conduct}: Population = Dutch PH research; Intervention = ML methods; Comparison = not applicable; Outcome = AB identification, discussion, and reporting across the three categories; Context = peer-reviewed Dutch PH+ML studies published between 2021–2025. Beyond the SLR, we introduce the ACAR framework to guide AF integration across the PH+ML research lifecycle.

\subsection{Inclusion and Exclusion Criteria}

Studies were included if they met all of the following:

\begin{itemize}
    \item \textbf{Topic Relevance}: Addressed PH topics such as population health, disease prevention, epidemiology, surveillance, or population-level interventions.
    \item \textbf{ML Application}: Trained ML models for prediction, classification, or related tasks.
    \item \textbf{Dutch Context}: Used Dutch data or included at least one author (first, second, or last) affiliated with a Netherlands-based institution.
    \item \textbf{Publication Type}: Peer-reviewed articles only.
    \item \textbf{Language}: English.
    \item \textbf{Publication Date}: January 2021 — February 2025.
\end{itemize}

Studies were excluded if any of the following applied:

\begin{itemize}
    \item \textbf{Clinical/Operational Focus}: Centered on clinical decision-making, biomedical mechanisms, or healthcare management (e.g., hospital operations, cost optimization) without clear PH relevance.
    \item \textbf{No Valid ML Use}: Misused ML terms (e.g., called regression or feature selection ML) or did not train models.
    \item \textbf{Technology-Only Scope}: Focused solely on technical development unrelated to PH outcomes (e.g., signal processing, device engineering).
\end{itemize}

\begin{figure}[t]
\centering
\includegraphics[width=\columnwidth]{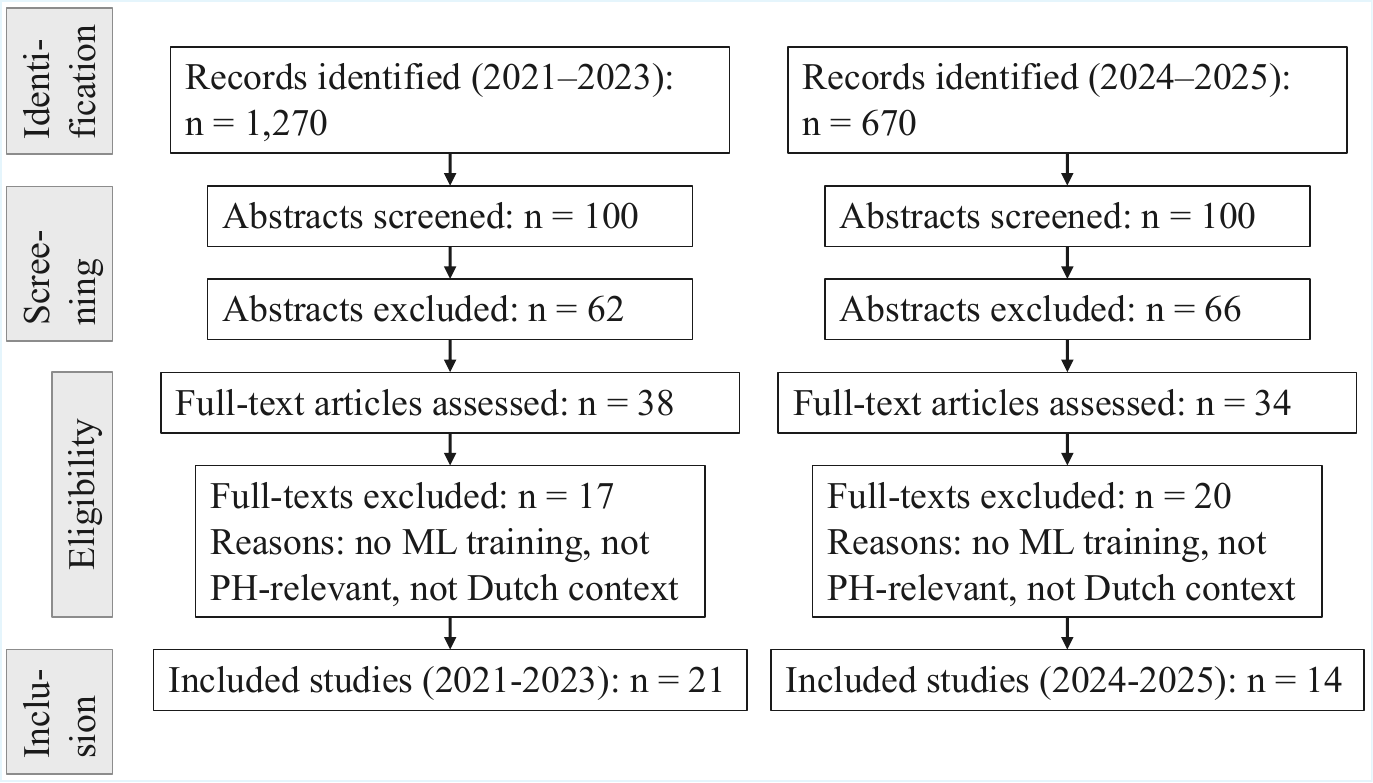}
\caption{PRISMA flow diagram of study selection process.}
\label{fig:prisma}
\end{figure}

When PH relevance was unclear, full-text screening assessed if the study’s objectives, population, outcomes, or implications supported population-level insights, policy decisions, or PH service delivery. Studies were included if they met our operational definition of PH+ML: ML methods relevant to population-level outcomes or potential to inform PH policy, surveillance, or interventions. The inclusion process is summarized in the PRISMA diagram (Figure~\ref{fig:prisma}).

\subsection{Search Strategy and Screening Process}

This review initially focused on AB in PH+ML research by GGD Amsterdam, the city’s municipal health service, given its policy role, robust data infrastructure, and diverse population. By 2030, an estimated 55.7\% of Amsterdam’s residents will have a non-Dutch background, including sizable Surinamese, Turkish, Moroccan, and Indonesian communities. These groups bear disproportionately high burdens of chronic diseases—especially diabetes, cardiovascular conditions, and mental health disorders—compared to ethnic Dutch populations \citep{ikram2014disease}. Combined with suboptimal cultural responsiveness in mental healthcare, Amsterdam is a salient case for AB in ADM \citep{ilozumba2022ethnic}. Its diversity and PH infrastructure provide a strong context to examine how ML interacts with population heterogeneity and equity outcomes \citep{essink2013interpreting}.

A manual review of GGD Amsterdam’s archive in December 2023 for peer-reviewed studies (2021–2023) mentioning ML found no eligible publications. Follow-up keyword searches confirmed the absence of ML or AI usage, prompting a shift to a national-level review of Dutch PH+ML research.

Given the topic’s interdisciplinary nature, with studies scattered across medical, epidemiological, and technical venues not consistently indexed by PubMed or Scopus, Google Scholar was selected. Its broad coverage and relevance-based sorting were well suited to capture cross-disciplinary work. Since ML and PH are defined inconsistently across fields, this approach helped capture studies otherwise missed due to terminological or disciplinary silos.

\subsubsection{2021–2023 Search and Screening}

The structured search was conducted in December 2023 using the query:

\begin{quote}
\texttt{("machine learning" AND "Public Health" AND "Netherlands") AND ("Amsterdam UMC" OR RIVM OR "Public Health Service")}
\end{quote}

The search returned 1,270 results, sorted by relevance. Abstract-level screening was extended to the first 200 records; however, no additional eligible studies were identified in abstracts 101–200, limiting our review to the top 100 abstracts. Screening followed a two-stage protocol: (1) abstract-level screening against inclusion criteria, and (2) full-text review for shortlisted studies. Ultimately, 21 studies published between 2021 and 2023 met all inclusion criteria.

\subsubsection{2024–2025 Updated Search and Screening}

To assess whether AF and AB considerations in Dutch PH+ML research have progressed, particularly amid the rise of generative AI tools such as large language models (LLMs), a follow-up search was conducted in February 2025. The same query and screening criteria were used to ensure consistency with the 2021–2023 review.

The search returned 670 results. Again, the first 100 abstracts were screened, with shortlisted studies reviewed in full. Fourteen studies met all inclusion criteria and were added to the sample.

In both search phases, studies were excluded if they lacked model training—such as those using ML methods only for feature selection, post hoc analyses, or variable importance without testing on unseen data. Also excluded were studies proposing ML methods without implementation, protocol studies with unrealized ML components, gray literature, theses, reviews, and non-peer-reviewed documents to ensure inclusion of genuine PH+ML applications.

All records were tracked in a structured spreadsheet, with detailed documentation of inclusion decisions.

\subsection{RABAT: Risk of Algorithmic Bias Assessment Tool}

To assess how AB is identified, discussed, and reported in Dutch PH+ML research, we developed RABAT, which selectively adapts elements from three frameworks. These were chosen for their credibility and complementary focus on people, processes, data, and methods, all of which are critical for an integral assessment of AB in PH+ML research (see Appendix Table~\ref{tab:rabat-mapping} for source-question mapping). RABAT adds novelty through its focus on PH+ML, structured scoring logic, and emphasis on AB and subgroup risks. First, the Cochrane Risk of Bias Tool identifies systematic errors in randomized trials, assessing domains like selection, performance, detection, and reporting bias—essential for evaluating study validity and transparency in health research \citep{higgins2011cochrane}. Second, PROBAST (2019) provides structured criteria for assessing bias in prediction model studies, focusing on how predictors are defined, selected, and handled, which are key considerations in ML-based health modeling \citep{moons2019probast, wolff2019probast}. Third, Microsoft’s Responsible AI (MS RAI) checklist emphasizes reflecting on societal impacts, fairness framing, and stakeholder harms throughout the ML lifecycle \citep{madaio2020co}. We derived three categories from this integration: \textit{extent of bias discussion}, \textit{identification of subgroups at risk}, and \textit{reporting transparency}. These categories reflect a PH perspective by placing people at the center of bias discussion, subgroup risk, and research reporting. For example, the societal impact question (Q4) draws from MS RAI checklist's prompts about the system’s role, affected stakeholders, and benefit–harm trade-offs. The ten resulting questions reflect our guiding research question. A condensed version of RABAT appears in Table~\ref{tab:rabat-condensed}; full item wording and grading criteria are provided in Appendix Tables~\ref{tab:rabat-cat1}--\ref{tab:rabat-grading}.

\begin{table}[t]
\centering
{\fontsize{9pt}{11pt}\selectfont
\begin{tabular}{p{0.04\linewidth} p{0.86\linewidth}}
\toprule
\textbf{Q\#} & \textbf{Condensed RABAT Questionnaire} \\
\midrule
Q1 & Is data bias discussed (e.g., representativeness)? \\
Q2 & Is model bias discussed (e.g., performance disparities)? \\
Q3 & Is bias framed in ML fairness terms? \\
Q4 & Are potential societal impacts discussed? \\
Q5 & Are at-risk subgroups identified? \\
Q6 & Are data sampling, inclusions, and exclusions described? \\
Q7 & Are sensitive attributes reported? \\
Q8 & Is bias described in sufficient length and structure? \\
Q9 & Is mitigation of fairness-related harms reported? \\
Q10 & Is meaningful informed consent discussed? \\
\bottomrule
\end{tabular}
}
\caption{Condensed version of the RABAT questionnaire.}
\label{tab:rabat-condensed}
\end{table}

\textbf{Scoring and Evaluation Process.}
Two reviewers independently applied RABAT to all studies from the 2021–2023 and 2024–2025 review phases. Prior to full review, the rubric was pilot-tested on a sample of papers to refine definitions and eliminate ambiguity. Discrepancies were discussed to finalize the scoring guide. Each RABAT item was scored on a four-point scale: 0 = absent, 1 = minimal, 2 = moderate, and 3 = extensive; higher scores indicate stronger performance on AB-related reporting. In addition, mean RABAT scores were classified into \textit{Low Risk}, \textit{Some Concerns}, or \textit{High Risk} using calibrated thresholds (\textless{}0.75 = High Risk; 0.75–1.5 = Some Concerns; \textgreater{}1.5 = Low Risk), adapted from PROBAST principles and adjusted for the observed left-skewed score distribution. Percentage agreement between reviewers was calculated per RABAT item (excluding \texttt{NA}). Agreement ranged from 71.4\% to 88.6\%. Highest rates were for \textit{ML fairness} (Q3, 88.6\%), \textit{sensitive attributes} (Q7), and \textit{harm transparency} (Q9), both 85.7\%. \textit{Societal impact} (Q4) reached 80.0\%, while \textit{model bias} (Q2), \textit{subgroups at risk} (Q5), and \textit{informed consent} (Q10) were near 77\%. \textit{Sampling and missing data} (Q6) had 74.3\%; the lowest agreement was for \textit{data bias} (Q1) and \textit{bias articulation} (Q8), both 71.4\%. These results reflect moderate to substantial agreement, strongest for clearly defined items. RABAT also served as a quality appraisal tool, replacing standard SLR checklists \citep{carrera2022conduct}, assessing presence and depth of AB reporting to evaluate transparency and rigor. Scores were logged in a spreadsheet; full rubric and examples are available on request.

\subsection{ACAR: Awareness, Conceptualization, Application, and Reporting} 
To synthesize our findings and support practical translation, we developed a four-stage framework adapted from Design Thinking. Design Thinking offers a flexible, problem-oriented approach widely used in applied fields to guide iterative learning, user-centered design, and structured reflection \citep{brown2009change}. Though rooted in design and engineering, its focus on awareness, iterative conceptualization, stakeholder relevance, and transparent problem-solving aligns with the needs of AB-aware ML. While general frameworks offer lifecycle guidance, ACAR adapts these principles to the PH+ML domain, tailoring its stages to domain-specific concerns such as subgroup relevance, societal impact, and context-aware reporting. It emphasizes AB identification, fairness conceptualization, method application, and transparent reporting as sequential stages in the research workflow (see Appendix Table~\ref{tab:dt_acar_mapping}). We define each of the four ACAR stages as follows:

\begin{enumerate}
    \item \textbf{Awareness:} Recognize that AB may emerge from data, model design, or social context. Reflect on whether fairness is relevant, who might be affected, and how broader societal impacts could arise.
    
    \item \textbf{Conceptualization:} Define AB, fairness, and subgroup risks in relation to research objectives and methodology. Frame these concepts early to establish a clear fairness lens.
    
    \item \textbf{Application:} Implement strategies to address AB in data and modeling workflows, including sampling decisions, subgroup testing, and bias mitigation techniques, whether experimental or in production.
    
    \item \textbf{Reporting:} Clearly communicate how AB risks and fairness considerations were addressed, including structured bias discussions, subgroup findings, transparency about limitations, and ethical elements like consent.
\end{enumerate}

\section{Results}

\subsection{Overview of Included Studies}

This review includes 35 peer-reviewed PH+ML articles published between 2021 and 2025 that met all inclusion criteria (Appendix Table~\ref{tab:included-studies}). The sample comprised nine studies from 2021, seven from 2022, five from 2023, twelve from 2024, and two from early 2025, reflecting a sharp increase in publications from 2024. Covered domains included infectious disease surveillance, mental health, behavioral prediction, environmental exposure, and chronic disease modeling.

Most studies applied classification (n = 21) or regression (n = 9); a few used time-series forecasting (n = 3) or survival analysis (n = 1). Common methods included Random Forest, Logistic Regression, and XGBoost, often combined with Support Vector Machines, LASSO, or deep learning models. Ensemble and hybrid approaches were also used. Performance metrics were consistently reported, with frequent use of area under the curve (AUC), sensitivity, specificity, accuracy, precision, and F1 score; fewer studies included calibration, confidence intervals, or explainability metrics such as SHapley Additive exPlanations (SHAP).

Dataset sizes ranged from small (e.g., 134 proteomics patients, 253 helpline transcripts) to large cohorts such as the Public Health Monitor (PHM) (244,557 individuals), national emergency department (ED) admissions for acute coronary syndrome (214,953 patients, 2010–2017), and the STIZON registry (over one million primary care records curated for research use). All studies used Dutch data or involved Dutch-affiliated authors, often through cross-sector collaborations spanning public health, clinical, and academic institutions.

Most studies addressed ethical or regulatory compliance, citing approvals from Dutch Medical Ethics Committees, GDPR waivers, or the Declaration of Helsinki. A minority lacked formal ethics or consent statements. Human oversight was typically reported through expert input or annotation. Real-world deployment was rare: most models were retrospective, with only two used in clinical or PH systems. Several proposed deployment but remained in development. The split sample design showed fairness practices were largely absent across both periods reviewed.

As described in Section~\ref{sec:methods}, all studies were evaluated using the ten-item RABAT framework, with each item scored from 0 (absent) to 3 (extensive). The following subsections present findings by category.

\begin{table}[ht]
\centering
{\fontsize{9pt}{11pt}\selectfont
\begin{tabular}{p{0.49\linewidth}|p{0.09\linewidth}|p{0.27\linewidth}}
\hline
\textbf{RABAT Question} & \textbf{Mean Score} & \textbf{Risk Level} \\
\hline
Q1. Data bias                        & 0.74 & High Risk \\
Q2. Model bias                       & 0.54 & High Risk \\
Q3. ML fairness                      & 0.06 & High Risk \\
Q4. Societal impact                  & 0.59 & High Risk \\
Q5. Subgroups at risk               & 0.37 & High Risk \\
Q6. Sampling \& missing data        & 1.64 & Low Risk \\
Q7. Sensitive attributes             & 0.07 & High Risk \\
Q8. Bias articulation                & 0.60 & High Risk \\
Q9. Harm transparency                & 0.13 & High Risk \\
Q10. Informed consent                & 0.96 & Some Concerns \\
\hline
\end{tabular}
}
\caption{Mean RABAT scores classified as Low Risk, Some Concerns, or High Risk using calibrated thresholds (\textless{}0.75, 0.75–1.5, \textgreater{}1.5) adapted for left-skewed distributions.}
\label{tab:rabat_summary}
\end{table}
\subsection{Category 1: Extent of Bias Discussion}

\subsubsection*{Q1: Data Bias}

The median score was 1.0 (IQR: 0.5--1.0), with a mean of 0.74 (SD = 0.46), classified as \textit{High Risk}. Six studies (17\%) did not mention data bias; 27 (77\%) engaged only minimally. Two studies (6\%) reached a moderate level; none were rated extensive. Most discussions were indirect---referring to class imbalance, missing data, or non-representativeness---without explicitly framing these as sources of bias. Common issues included subgroup over-representation, incomplete linkage, or limited access to key variables (e.g., age, sex, socioeconomic status). Some papers addressed data quality, stratification, or exclusions, but rarely linked these to biased outcomes. Only one study described a mitigation strategy; none conducted structured assessments. Mentions of provenance, measurement error, or distributional skew were largely absent.

\subsubsection*{Q2: Model Bias}

The median score was 0.5 (IQR: 0.0--1.0), with a mean of 0.54 (SD = 0.48), classified as \textit{High Risk}. Thirteen studies (37\%) did not address model bias; 21 (60\%) engaged only minimally, typically noting overfitting, performance variation, or majority-class favoring. One study (3\%) scored moderate; none were extensive. Discussions were brief and framed as technical concerns. Some cited uncertainty from small datasets, regional misclassification, or modest predictive performance, but without linking these to structural risks. A few mentioned subgroup underperformance or the need for validation, but none reported disaggregated error, architectural bias, or AB mitigation efforts. No study conducted formal audits or addressed AF risks in ADM.

\subsubsection*{Q3: ML Fairness}

The median score was 0.0 (IQR: 0.0--0.0), with a mean of 0.06 (SD = 0.16), classified as \textit{High Risk}. Thirty-one studies (89\%) did not mention fairness; the remaining four (11\%) engaged only minimally. None were rated moderate or extensive. Fairness was largely absent as an analytic or ethical concept. Indirect mentions referred to generalizability, overfitting, or geographic variation, without linking to protected attributes or subgroup harm. No study defined fairness, used fairness metrics, or applied fairness-aware techniques. Even when model bias was noted, its potential impact on specific populations was not explored. ML fairness remained outside the conceptual and methodological scope of nearly all studies.

\subsubsection*{Q4: Societal Impact}

The median score was 0.5 (IQR: 0.0--1.0), with a mean of 0.59 (SD = 0.51), classified as \textit{High Risk}. Twelve studies (34\%) did not mention societal impacts; 22 (63\%) engaged only minimally, typically referencing clinical utility, PH relevance, or policy applications. One study (3\%) scored moderate; none were extensive. When discussed, societal impacts centered on potential benefits (e.g., screening, efficiency, decision support), while harms, trade-offs, and stakeholder-specific effects were rarely addressed. Some noted implications for workflows or planning, but equity, access, and unintended consequences were seldom considered. No study examined differential impacts on vulnerable groups or broader ethical dimensions of ADM.

\subsection{Category 2: Identification of Subgroups at Risk}

\subsubsection*{Q5: Subgroups at Risk}

The median score was 0.0 (IQR: 0.0--0.5), with a mean of 0.37 (SD = 0.52), classified as \textit{High Risk}. Twenty studies (57\%) did not identify any at-risk subgroups; thirteen (37\%) engaged only minimally, typically referencing broad demographic categories (e.g., age, sex, comorbidities) without analyzing differential model performance. Two studies (6\%) scored moderate; none were extensive. Mentions of subgroup risks were often indirect or speculative, such as noting exclusion of severely affected patients, limited data for specific groups (e.g., immunocompromised individuals), or demographic imbalances. Several studies listed protected attributes or used stratified analyses, but without linking these to fairness. No study systematically assessed whether ML models produced disparate outcomes across subpopulations or investigated sources of differential error or harm.

\subsubsection*{Q6: Sampling and Missing Data}

This item received the highest scores across all RABAT questions, with a median of 2.0 (IQR: 1.0--2.0) and a mean of 1.64 (SD = 0.80), classified as \textit{Low Risk}. Three studies (9\%) included no relevant content; nine (26\%) provided only minimal information. The remaining 23 (66\%) described sampling and missing data handling at moderate or extensive levels. Most clearly identified datasets and eligibility criteria, often reporting variable distributions by sex, age, or location. Several justified exclusions or discussed implications of missing data. Common imputation methods included median substitution and multiple imputation (e.g., MICE), though few assessed their impact on model performance. Fairness considerations, such as whether missingness or exclusions disproportionately affected specific groups, were absent. Even in detailed methodological reporting, representation bias and subgroup exclusions were not discussed in fairness terms.

\subsubsection*{Q7: Sensitive Attributes}

The median score was 0.0 (IQR: 0.0--0.0), with a mean of 0.07 (SD = 0.18), classified as \textit{High Risk}. Thirty studies (86\%) did not mention sensitive or fairness-relevant attributes such as race, ethnicity, disability, or migration background. Five (14\%) engaged only minimally, typically listing variables like sex or age in tables or covariates, without analyzing model output variation across groups. No studies scored moderate or extensive. While many studies collected sociodemographic data or stratified by sex, none assessed disproportionate model errors, subgroup representation, or selection risks. As with Q6, no study problematized links between sensitive attributes and potential harms from ADM.

\subsection{Category 3: Reporting Transparency}

\subsubsection*{Q8: Bias Articulation}

The median score was 0.5 (IQR: 0.0--1.0), with a mean of 0.60 (SD = 0.48), classified as \textit{High Risk}. Eleven studies (31\%) did not discuss bias risks; twenty-two (63\%) engaged only minimally, typically referencing imbalanced data, generalizability, or limited data quality. Two studies (6\%) provided a moderate discussion, though engagement was often scattered or implicit. Most framed bias in technical terms, such as overfitting, missing data, or performance degradation across sites. Few treated bias as a systemic risk requiring mitigation, and none offered a dedicated section or structured analysis. Limitations were often described as general uncertainty or data quality issues, not as fairness-relevant concerns. No study examined how AB might propagate through the pipeline or affect subgroups differentially.

\subsubsection*{Q9: Harm Transparency}

The median score was 0.0 (IQR: 0.0--0.0), with a mean of 0.13 (SD = 0.28), classified as \textit{High Risk}. Twenty-eight studies (80\%) made no mention of fairness-related harms or mitigation. Seven (20\%) acknowledged bias mitigation techniques—such as SMOTE, class weighting, or exclusion of correlated variables—but did not assess whether these reduced fairness risks. No study reported how harms were identified, assessed, or mitigated across the ML lifecycle. Bias was typically framed as a technical challenge affecting performance, not a source of downstream impact on populations. Even in PH or clinical settings, subgroup harms were not considered. None included audits, fairness checks, or mitigation justifications grounded in ethical or equity frameworks.

\subsubsection*{Q10: Informed Consent}

The median score was 1.0 (IQR: 0.0--1.875), with a mean of 0.96 (SD = 0.86), classified as \textit{Some Concerns}. Eight studies (23\%) did not mention informed consent. Ten studies (29\%) met the minimal threshold, typically through brief statements of ethics approval or consent, or by citing waivers without elaboration on scope, process, or participant understanding. Six studies (17\%) provided moderate reporting, including ethics approvals and confirmation of written or institutional consent, though none addressed risks of re-identification or model reuse. Only two studies (6\%) scored extensive. Across the sample, informed consent was generally treated as a procedural requirement rather than an ethical concern, and no study engaged with the challenges of meaningful consent in ADM contexts.

\begin{figure*}[t]
  \centering
  \includegraphics[width=\textwidth]{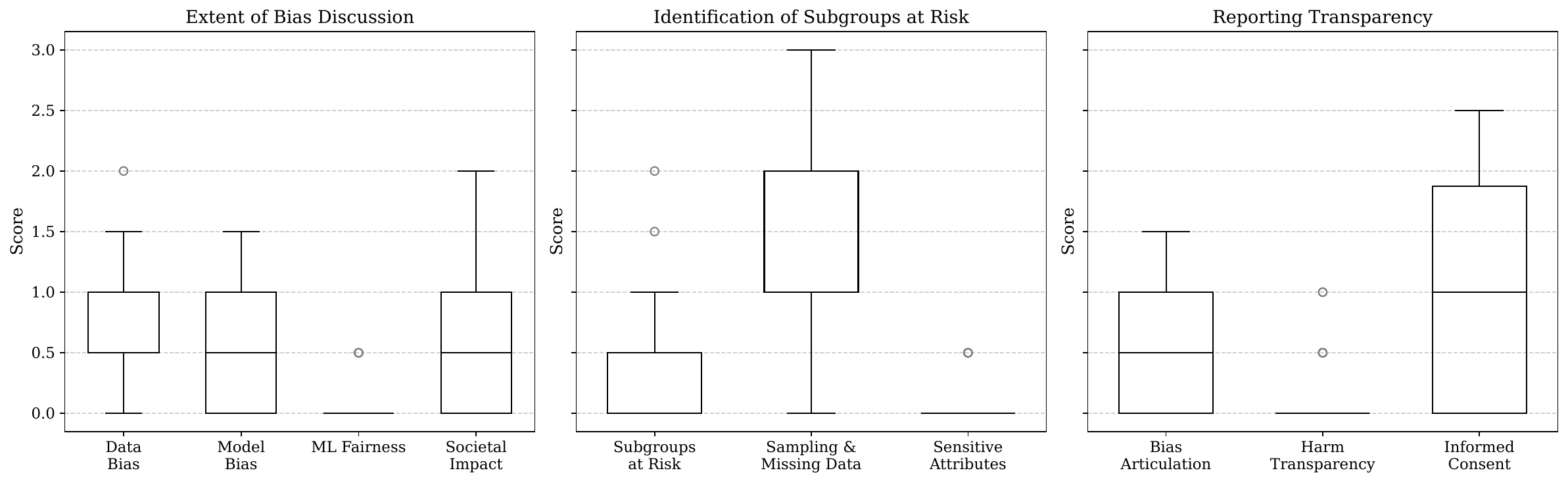}
  \caption{RABAT score distribution across ten questions, grouped by category. While some AB reporting aspects (e.g., \textit{sampling and missing data}, \textit{informed consent}) show moderate coverage, fairness-specific items (e.g., \textit{ML fairness}, \textit{sensitive attributes}, \textit{harm transparency}) score consistently low, revealing critical gaps in Dutch PH+ML research.}
  \label{fig:boxplot_rabat}
\end{figure*}

\subsection{Overall Scoring Summary}

Of the ten RABAT items, only one—\textit{Sampling and Missing Data} (Q6)—had a mean score above 1.0 and a median of 2.0 (IQR: 1.0–2.0), indicating generally moderate, occasionally extensive, reporting. In contrast, eight items had mean scores below 0.75 and were classified as \textit{High Risk}, including three—\textit{ML Fairness} (Q3), \textit{Sensitive Attributes} (Q7), and \textit{Harm Transparency} (Q9)—with means under 0.15, reflecting near-total absence of fairness-specific reporting. Figure~\ref{fig:boxplot_rabat} shows the score distributions. Items tied to conventional epidemiologic reporting, such as Q6 and \textit{Informed Consent} (Q10), had higher medians and visibly wider score ranges. By contrast, fairness-oriented items like Q3, Q7, and Q9 were heavily skewed, with most studies scoring zero and minimal spread. \textit{ML Fairness}, for example, received no scores above 1 (minimal level). These contrasts highlight a clear asymmetry between methodological rigor and fairness-aware reporting in PH+ML research.

Table~\ref{tab:rabat_summary} summarizes score distributions by question. Only Q6 was classified as \textit{Low Risk}, with two-thirds of studies providing moderate-level reporting. Q10 fell in the \textit{Some Concerns} range, with several studies reporting moderately but many offering limited detail. The remaining eight items were classified as \textit{High Risk}, reflecting low mean scores and frequent absence of relevant content. Notably, the lowest mean scores were for fairness-specific items, indicating systemic neglect in this area.

\subsection{ACAR Framework Adaptation and Guidance}

To align evidence with practice, we mapped RABAT questions to the ACAR stages and analyzed associated patterns. In the \textit{Awareness} stage, most studies mentioned data (Q1) or model limitations (Q2), but rarely linked these to structural bias or societal risks (Q4). Only one study discussed a mitigation strategy for data bias, and just one reached moderate-level discussion of societal impact, with none rated extensive. For \textit{Conceptualization}, fairness was absent in 89\% of studies (Q3), and subgroup vulnerabilities (Q5) were largely unaddressed, with only two studies scoring moderate. In the \textit{Application} stage, while sampling and missing data (Q6) were consistently reported (66\% moderate/extensive), sensitive attributes (Q7) and subgroup testing or mitigation (Q9) were almost entirely absent—both had mean scores below 0.15. Finally, in \textit{Reporting}, some studies articulated bias (Q8) or mentioned consent (Q10), but few addressed fairness-related harms (Q9), and no study included structured audits or mitigation justifications. These gaps motivate our proposed guiding questions to support fairness-aware practice across the PH+ML workflow.

\subsubsection{Key Gaps and Weaknesses}

The most pronounced deficiencies appeared at the \textit{Conceptualization}, \textit{Application}, and \textit{Reporting} stages. Fairness was rarely defined (Q3), subgroup risks were largely unexamined (Q5), and no study applied fairness metrics (Q3) or conducted structured mitigation and transparent harm reporting (Q9). Sensitive attributes (Q7) were routinely omitted, and structured harm assessments were absent or superficial. Reporting focused primarily on technical limitations, with minimal articulation of AB (Q8) and limited detail on consent practices (Q10). These patterns reflect a pervasive lack of fairness-oriented engagement across the PH+ML workflow.

\subsubsection{Guiding Questions for PH+ML Researchers}

To support structured reflection and fairness integration across the PH+ML research lifecycle, we propose a set of guiding questions organized by ACAR stage. Designed to complement the RABAT assessment, these prompts help translate fairness considerations into concrete actions across study design, model development, and reporting. The questions are general by design and intentionally crafted to be accessible to interdisciplinary teams, reflecting our finding that PH+ML research is often conducted by collaborators with diverse disciplinary backgrounds, and varying levels of technical expertise. Table~\ref{tab:acar_guiding_questions} presents guiding questions to help researchers operationalize fairness in PH+ML.

\begin{table*}[t]
\centering
\begin{tabular}{@{}p{3cm}p{14cm}@{}}
\toprule
\textbf{ACAR Stage} & \textbf{Guiding Questions} \\
\midrule
\textit{Awareness} &
\textbf{Bias Sources:} Have you identified potential bias from data, models, or population differences? \newline
\textbf{Subgroup Impact:} Could your model perform differently across population subgroups? \newline
\textbf{Equity Risks:} Could your model worsen existing health inequities or exclude vulnerable groups? \newline
\textbf{Transparency:} Are risks and societal impacts clearly described? \\
\addlinespace
\textit{Conceptualization} &
\textbf{Fairness Framing:} Have you defined fairness in relation to your population and goals? \newline
\textbf{At-Risk Subgroups:} Have you identified groups that may be disproportionately affected? \newline
\textbf{Bias Mechanisms:} Have you considered how the model might produce disparities? \newline
\textbf{Subgroup Relevance:} Are subgroup definitions grounded in public health disparities? \\
\addlinespace
\textit{Application} &
\textbf{Sampling Fairness:} Have you assessed whether missing data or exclusions affect certain groups? \newline
\textbf{Sensitive Attributes:} Do key traits like sex, age, or origin influence predictions? \newline
\textbf{Subgroup Testing:} Have you tested for unequal performance across key groups? \newline
\textbf{Mitigation Actions:} Have you applied and evaluated bias reduction strategies? \\
\addlinespace
\textit{Reporting} &
\textbf{Bias Reporting:} Are bias sources and their subgroup effects clearly described? \newline
\textbf{Disparity Disclosure:} Have you reported group disparities and their implications? \newline
\textbf{Mitigation Reporting:} Have you reported bias mitigation actions and their outcomes? \newline
\textbf{Consent Clarity:} Is consent clearly reported, including data use and fairness risks? \\
\bottomrule
\end{tabular}
\caption{Guiding questions for AB-aware PH+ML research, labeled for integration across study design, modeling, and reporting.}
\label{tab:acar_guiding_questions}
\end{table*}

\section{Discussion}
Understanding how AB risk is discussed and reported in PH+ML is essential for evaluating real-world harms and guiding equitable system design. Although ML adoption in PH is accelerating, the field lacks consistent standards for identifying which AB risks matter, to whom, and how they should be reported. This review addresses that gap by applying a structured lens to assess AB-related transparency and synthesizing reporting patterns in recent Dutch PH+ML studies. Gaps such as the lack of fairness metrics or subgroup error analysis may reflect limited methodological guidance, a focus on technical performance, or unfamiliarity with fairness-aware tools. To examine these patterns, we systematically evaluated 35 studies using RABAT, a ten-item tool assessing how AB risk is addressed in PH+ML research. Questions covered three areas: extent of bias discussion, identification of subgroups at risk, and reporting transparency, scored from 0 (absent) to 3 (extensive). This revealed consistent gaps—particularly in fairness framing, subgroup focus, and harm reporting—alongside variation in overall reporting quality. In contrast, data sampling and informed consent were more commonly addressed, likely reflecting established biomedical norms. These insights informed the development of the ACAR framework to support fairness integration in PH+ML workflows.

\subsection{Observed Insights}

Final RABAT scores revealed substantial variability in how studies identified, discussed, and reported the risk of AB. Figure~\ref{fig:boxplot_rabat} helps us visualize question-level score distributions, highlighting two distinct reporting patterns. First, items aligned with conventional epidemiological practices—such as \textit{Sampling and Missing Data} (Q6) and \textit{Informed Consent} (Q10)—showed higher median scores and greater dispersion across the scale, indicating both engagement and heterogeneity in reporting practices. In particular, Q6 displayed the highest central tendency (median = 2.0) and interquartile range, reflecting wide uptake of standard reporting norms. These patterns are echoed in Table~\ref{tab:rabat_summary}, where only Q6 was classified as \textit{Low Risk} and Q10 as \textit{Some Concerns}. In contrast, fairness-specific items—including \textit{ML Fairness} (Q3), \textit{Sensitive Attributes} (Q7), and \textit{Harm Transparency} (Q9)—were clustered near zero, with minimal variance, suggesting widespread omission rather than inconsistency. Correspondingly, these three items received mean scores below 0.15 and were classified as \textit{High Risk}. Bias-related items such as \textit{Data Bias} (Q1), \textit{Model Bias} (Q2), and \textit{Subgroups at Risk} (Q5) fell in between, with low to moderate engagement and scores concentrated around minimal levels. The shape and spread of these boxplots underscore a key observation: while epidemiologic-reporting rigor was often maintained, fairness considerations remained underdeveloped, sporadic, or absent altogether. Even when sensitive features were collected or subgroup labels applied, their implications for fairness or subgroup harm were seldom analyzed. No study applied fairness metrics or conducted disaggregated performance evaluations, underscoring the persistent absence of structural AB risk assessment in routine Dutch PH+ML reporting.

\subsection{Actionable Recommendations for PH+ML Researchers}

Derived from RABAT findings, these recommendations support consistent AB-aware practices in PH+ML research.

\begin{enumerate}
\item \textbf{State and Justify AB Risks.} Identify sources of data and model bias (e.g., overrepresentation, missingness) and explain their potential effects.

\item \textbf{Define Fairness Early.} Provide a study-relevant definition of AF (e.g., group or individual fairness) and specify its relevance.

\item \textbf{Identify Subgroups Proactively.} Note any populations likely to face higher AB risk, even if not analyzed separately.

\item \textbf{Test by Subgroup.} Disaggregate performance by subgroup when feasible, and report disparities clearly.

\item \textbf{Audit for Harms.} Assess downstream harms or unintended effects, or explicitly acknowledge when not assessed.

\item \textbf{Report Bias Transparently.} Include a dedicated section on AB sources, subgroup risks, and AF limitations.

\item \textbf{Use the ACAR Guide.} Refer to ACAR during planning and reporting to ensure AF is considered throughout.
\end{enumerate}

\subsection{Limitations}
This review has some limitations. Although Dutch-language and institutional sources were manually searched, all included studies were from Google Scholar and English-language peer-reviewed sources, which may have limited the scope of the review, excluding Dutch-language, gray, or discipline-specific literature. The use of a single reviewer for initial screening may have introduced selection bias, though a structured two-stage protocol was followed. While piloting and inter-reviewer checks supported scoring consistency, RABAT’s reliability, generalizability beyond this context, and usability by non-expert reviewers have not yet been assessed. Only two reviewers per paper conducted scoring, which may limit reproducibility, though reviewer alignment was essential due to the complexity and time demands of RABAT. As the review aimed to identify reporting patterns rather than test hypotheses, no statistical analyses were conducted. The ACAR framework also remains theoretical, as it has not yet been applied in real-world PH+ML settings. Finally, findings reflect the Dutch context and may, for example, not generalize to low-resource or alternative governance settings.
Importantly, this review did not aim to exhaustively catalog all Dutch PH+ML studies. Nonetheless, the purposive sample of 35 studies was sufficient to reveal consistent reporting gaps, despite potential selection bias from manual screening. Through an SLR and structured RABAT assessment, we identified representative patterns to inform the ACAR framework—proposed here as a conceptual guide, not yet validated in real-world settings, and positioned as part of a broader mixed-methods strategy to embed fairness-aware practice in PH+ML research.

\subsection{Closing the Gap}

Our review confirms and extends prior findings on gaps in fairness-aware practices in PH+ML research. Earlier work has highlighted the limited use of AF metrics, subgroup evaluation, and harm transparency in these studies \citep{mhasawade2021machine, delgado2022bias, chen2024unmasking, raza2024exploring}. These reviews also identify a deeper conceptual gap: AF is often reduced to technical performance, with limited attention to structural determinants or downstream harms \citep{wesson2022risks, chin2023guiding, char2020identifying}.
Although \citet{rajkomar2018ensuring}, \citet{vollmer2020machine}, and others such as \citet{fletcher2021addressing}, \citet{char2020identifying}, \citet{chin2023guiding}, \citet{morgenstern2020predicting}, and \citet{raza2023connecting} offer principles, checklists, or guiding criteria to address AF, most remain either high-level, clinical in scope, or insufficiently tailored to the structural and interdisciplinary challenges of PH+ML.

We respond to this gap by introducing ACAR, a PH-oriented, stage-based framework grounded in our empirical findings. Unlike prior approaches, ACAR is directly derived from RABAT-assessed reporting gaps and provides targeted guidance for fairness-aware ML design, evaluation, and reporting in PH contexts. As detailed in Table~\ref{tab:acar_guiding_questions}, the framework translates observed AB reporting gaps into actionable prompts, facilitating AF reflection throughout the PH+ML lifecycle. These guiding questions are intentionally designed for interdisciplinary teams, reflecting our observation that PH+ML research commonly involves collaborators with diverse expertise and backgrounds.

\subsection{Implications for Practice and Policy}

The lack of structured AB assessments in PH+ML research calls for systemic changes in education, research design, and governance. ACAR offers a practical set of questions to embed fairness reflection across PH+ML workflows. While not a technical tool, its staged structure helps diverse research teams identify AB risks, consider fairness, and improve reporting, enhancing transparency, replicability, and comparability. The questions were intentionally designed with a low barrier to adoption, allowing integration into existing projects regardless of ML maturity. ACAR complements existing protocols without requiring significant changes.
However, institutionalizing these practices may face obstacles such as limited awareness, weak incentives, and unclear responsibility for fairness within teams. Targeted training, evaluation mandates, and clearer roles may help overcome these barriers. Educational programs can include fairness case studies and interdisciplinary training to build capacity for AB engagement through ACAR.
Policy measures such as the Dutch Algorithm Register and the CBS classification reform can support fairness reflection in PH+ML by enhancing algorithm transparency and improving equity in population classifications. Alignment with the EU AI Act and national AI strategies can further support AB reporting, though current efforts focus on sectors like law enforcement and healthcare rather than PH. Journals and funders can also promote shifts in norms by requiring AB discussion in publications and proposals. Ultimately, AF in PH+ML must become routine, not optional, in responsible research.

\subsection{Future Work}
We recommend (1) empirically validating ACAR in active PH+ML projects; (2) exploring researchers’ decision‐making around AB (e.g., interviews or surveys) to understand whether and why fairness considerations are addressed but not documented; (3) applying RABAT and ACAR in non‐Dutch contexts to test their applicability and identify any necessary context‐specific adaptations; (4) developing and testing training tools (e.g., workshops or online modules) to support ACAR use; (5) using collected metadata to explore additional patterns in AB reporting, including differences by venue or discipline; and (6) extending future reviews to include local, non-indexed, or non-English sources to better capture underrepresented work.

\section{Conclusion}

This study provides the first systematic assessment of AB reporting in Dutch PH+ML research. Applying the RABAT framework to 35 studies, we reveal that, despite strong technical rigor in data handling, fairness considerations—especially ML‐specific framing, sensitive‐attribute analysis, and transparent harm disclosure—are largely absent. Our ACAR framework offers a clear, actionable pathway to embed fairness at every stage of the ML lifecycle. We urge PH+ML researchers to adopt ACAR‐informed reporting practices and leverage national transparency mechanisms to ensure that future PH+ML applications advance health equity rather than exacerbate existing disparities.  

\section*{Acknowledgments}
This research was conducted at the Civic AI Lab (SIAS group), Informatics Institute, University of Amsterdam, in collaboration with the Municipality of Amsterdam and the Dutch Ministry of the Interior and Kingdom Relations.


\bibliography{aaai25}

\begin{thebibliography}{96}
\providecommand{\natexlab}[1]{#1}

\bibitem[{Altman and Saunders(1998)}]{altman1998credit}
Altman, E.~I.; and Saunders, A. 1998.
\newblock Credit Risk Measurement: Developments over the Last 20 Years.
\newblock \emph{Journal of Banking \& Finance}, 21(11-12): 1721--1742.

\bibitem[{Angwin et~al.(2016)Angwin, Larson, Mattu, and Kirchner}]{angwin2016machine}
Angwin, J.; Larson, J.; Mattu, S.; and Kirchner, L. 2016.
\newblock Machine Bias.
\newblock \url{https://www.propublica.org/article/machine-bias-risk-assessments-in-criminal-sentencing}.
\newblock ProPublica, May 23.

\bibitem[{Arnold and Pistilli(2012)}]{arnold2012course}
Arnold, K.~E.; and Pistilli, M.~D. 2012.
\newblock Course Signals at Purdue: Using Learning Analytics to Increase Student Success.
\newblock In \emph{Proceedings of the 2nd International Conference on Learning Analytics and Knowledge}, 267--270. Vancouver, BC, Canada: ACM.

\bibitem[{Aïvodji et~al.(2019)Aïvodji, Arai, Fortineau, Gambs, Hara, and Tapp}]{aivodji2019fairwashing}
Aïvodji, U.; Arai, H.; Fortineau, O.; Gambs, S.; Hara, S.; and Tapp, A. 2019.
\newblock Fairwashing: The Risk of Rationalization.
\newblock In \emph{Proceedings of the 36th International Conference on Machine Learning}, volume~97 of \emph{Proceedings of Machine Learning Research}, 161--170.

\bibitem[{Barocas, Hardt, and Narayanan(2023)}]{barocas-hardt-narayanan}
Barocas, S.; Hardt, M.; and Narayanan, A. 2023.
\newblock \emph{Fairness and Machine Learning: Limitations and Opportunities}.
\newblock MIT Press.
\newblock ISBN 9780262048613.

\bibitem[{Barocas and Selbst(2016)}]{Barocas_2016}
Barocas, S.; and Selbst, A.~D. 2016.
\newblock Big Data’s Disparate Impact.
\newblock \emph{SSRN Electronic Journal}.

\bibitem[{Bellamy et~al.(2019)Bellamy, Dey, Hind, Hoffman, Houde, Kannan, Lohia, Martino, Mehta, Mojsilović, Nagar, Ramamurthy, Richards, Saha, Sattigeri, Singh, Varshney, and Zhang}]{bellamy2019aif360}
Bellamy, R. K.~E.; Dey, K.; Hind, M.; Hoffman, S.~C.; Houde, S.; Kannan, K.; Lohia, P.; Martino, J.; Mehta, S.; Mojsilović, A.; Nagar, S.; Ramamurthy, K.~N.; Richards, J.; Saha, D.; Sattigeri, P.; Singh, M.; Varshney, K.~R.; and Zhang, Y. 2019.
\newblock {AI Fairness 360: An Extensible Toolkit for Detecting and Mitigating Algorithmic Bias}.
\newblock \emph{IBM Journal of Research and Development}, 63(4/5): 4:1--4:15.

\bibitem[{Benke and Benke(2018)}]{benke2018artificial}
Benke, K.; and Benke, G. 2018.
\newblock Artificial Intelligence and Big Data in Public Health.
\newblock \emph{International Journal of Environmental Research and Public Health}, 15(12): 2796.

\bibitem[{Binns(2018)}]{binns2018fairness}
Binns, R. 2018.
\newblock Fairness in Machine Learning: Lessons from Political Philosophy.
\newblock In Friedler, S.~A.; and Wilson, C., eds., \emph{Proceedings of the 2018 Conference on Fairness, Accountability and Transparency}, volume~81 of \emph{Proceedings of Machine Learning Research}, 149--159. PMLR.

\bibitem[{Bird et~al.(2020)Bird, Dudik, Edgar, Horn, Lutz, Milan, Sameki, Wallach, and Walker}]{bird2020fairlearn}
Bird, S.; Dudik, M.; Edgar, R.; Horn, B.; Lutz, R.; Milan, V.; Sameki, M.; Wallach, H.; and Walker, K. 2020.
\newblock Fairlearn: A Toolkit for Assessing and Improving Fairness in AI.
\newblock Technical Report MSR-TR-2020-32, Microsoft Research.

\bibitem[{Bossuyt et~al.(2015)Bossuyt, Reitsma, Bruns, Gatsonis, Glasziou, Irwig, Lijmer, Moher, Rennie, de~Vet, Kressel, Rifai, Golub, Altman, Hooft, Korevaar, and Cohen}]{Bossuyt_2015}
Bossuyt, P.~M.; Reitsma, J.~B.; Bruns, D.~E.; Gatsonis, C.~A.; Glasziou, P.~P.; Irwig, L.; Lijmer, J.~G.; Moher, D.; Rennie, D.; de~Vet, H.~C.; Kressel, H.~Y.; Rifai, N.; Golub, R.~M.; Altman, D.~G.; Hooft, L.; Korevaar, D.~A.; and Cohen, J.~F. 2015.
\newblock STARD 2015: An Updated List of Essential Items for Reporting Diagnostic Accuracy Studies.
\newblock \emph{Radiology}, 277(3): 826--832.

\bibitem[{Brown(2009)}]{brown2009change}
Brown, T. 2009.
\newblock \emph{Change by Design: How Design Thinking Creates New Alternatives for Business and Society}.
\newblock New York: HarperBusiness.
\newblock ISBN 9780061766084.

\bibitem[{Buolamwini and Gebru(2018)}]{buolamwini2018gender}
Buolamwini, J.; and Gebru, T. 2018.
\newblock Gender Shades: Intersectional Accuracy Disparities in Commercial Gender Classification.
\newblock In \emph{Proceedings of the 1st Conference on Fairness, Accountability and Transparency}, volume~81 of \emph{Proceedings of Machine Learning Research}, 77--91. PMLR.

\bibitem[{Burrell(2016)}]{burrell2016how}
Burrell, J. 2016.
\newblock How the machine ‘thinks’: Understanding opacity in machine learning algorithms.
\newblock \emph{Big Data \& Society}, 3(1): 1--12.

\bibitem[{Carrera-Rivera et~al.(2022)Carrera-Rivera, Ochoa, Larrinaga, and Lasa}]{carrera2022conduct}
Carrera-Rivera, A.; Ochoa, W.; Larrinaga, F.; and Lasa, G. 2022.
\newblock How-to conduct a systematic literature review: A quick guide for computer science research.
\newblock \emph{MethodsX}, 9: 101895.

\bibitem[{Caton and Haas(2024)}]{caton2024fairness}
Caton, S.; and Haas, C. 2024.
\newblock Fairness in Machine Learning: A Survey.
\newblock \emph{ACM Computing Surveys}, 56(7): 1--38.

\bibitem[{{Centraal Bureau voor de Statistiek (CBS)}(2022)}]{cbs2022}
{Centraal Bureau voor de Statistiek (CBS)}. 2022.
\newblock Migration Background No Longer Used as Standard Variable.
\newblock Accessed April 21, 2025.

\bibitem[{Chan et~al.(2013)Chan, Tetzlaff, Altman, Laupacis, Gøtzsche, Krleža-Jerić, Hróbjartsson, Mann, Dickersin, Berlin, Doré, Parulekar, Summerskill, Groves, Schulz, Sox, Rockhold, Rennie, and Moher}]{chan2013spirit}
Chan, A.-W.; Tetzlaff, J.~M.; Altman, D.~G.; Laupacis, A.; Gøtzsche, P.~C.; Krleža-Jerić, K.; Hróbjartsson, A.; Mann, H.; Dickersin, K.; Berlin, J.~A.; Doré, C.~J.; Parulekar, W.~R.; Summerskill, W. S.~M.; Groves, T.; Schulz, K.~F.; Sox, H.~C.; Rockhold, F.~W.; Rennie, D.; and Moher, D. 2013.
\newblock SPIRIT 2013 Statement: Defining Standard Protocol Items for Clinical Trials.
\newblock \emph{Annals of Internal Medicine}, 158(3): 200--207.

\bibitem[{Char, Abràmoff, and Feudtner(2020)}]{char2020identifying}
Char, D.~S.; Abràmoff, M.~D.; and Feudtner, C. 2020.
\newblock Identifying Ethical Considerations for Machine Learning Healthcare Applications.
\newblock \emph{The American Journal of Bioethics}, 20(11): 7--17.

\bibitem[{Chen et~al.(2024)Chen, Wang, Hong, Jiang, and Zhou}]{chen2024unmasking}
Chen, F.; Wang, L.; Hong, J.; Jiang, J.; and Zhou, L. 2024.
\newblock Unmasking Bias in Artificial Intelligence: A Systematic Review of Bias Detection and Mitigation Strategies in Electronic Health Record-Based Models.
\newblock \emph{Journal of the American Medical Informatics Association}, 31(5): 1172--1183.

\bibitem[{Chin et~al.(2023)Chin, Afsar-Manesh, Bierman, Chang, Colón-Rodríguez, Dullabh, Duran, Fair, Hernandez-Boussard, Hightower, Jain, Jordan, Konya, Moore, Moore, Rodriguez, Shaheen, Snyder, Srinivasan, Umscheid, and Ohno-Machado}]{chin2023guiding}
Chin, M.~H.; Afsar-Manesh, N.; Bierman, A.~S.; Chang, C.; Colón-Rodríguez, C.~J.; Dullabh, P.; Duran, D.~G.; Fair, M.; Hernandez-Boussard, T.; Hightower, M.; Jain, A.; Jordan, W.~B.; Konya, S.; Moore, R.~H.; Moore, T.~T.; Rodriguez, R.; Shaheen, G.; Snyder, L.~P.; Srinivasan, M.; Umscheid, C.~A.; and Ohno-Machado, L. 2023.
\newblock Guiding Principles to Address the Impact of Algorithm Bias on Racial and Ethnic Disparities in Health and Health Care.
\newblock \emph{JAMA Network Open}, 6(12): e2345050.

\bibitem[{Chouldechova(2017)}]{chouldechova2017fair}
Chouldechova, A. 2017.
\newblock Fair Prediction with Disparate Impact: A Study of Bias in Recidivism Prediction Instruments.
\newblock \emph{Big Data}, 5(2): 153--163.

\bibitem[{Collins et~al.(2024)Collins, Moons, Dhiman, Riley, Beam, Van~Calster, Ghassemi, Liu, Reitsma, van Smeden et~al.}]{collins2024tripodai}
Collins, G.~S.; Moons, K. G.~M.; Dhiman, P.; Riley, R.~D.; Beam, A.~L.; Van~Calster, B.; Ghassemi, M.; Liu, X.; Reitsma, J.~B.; van Smeden, M.; et~al. 2024.
\newblock {TRIPOD}+{AI} statement: updated guidance for reporting clinical prediction models that use regression or machine learning methods.
\newblock \emph{BMJ}, 385: e078378.

\bibitem[{Collins et~al.(2015)Collins, Reitsma, Altman, and Moons}]{collins2015tripod}
Collins, G.~S.; Reitsma, J.~B.; Altman, D.~G.; and Moons, K. G.~M. 2015.
\newblock Transparent Reporting of a Multivariable Prediction Model for Individual Prognosis or Diagnosis (TRIPOD): The TRIPOD Statement.
\newblock \emph{Circulation}, 131(2): 211--219.

\bibitem[{Corbett-Davies et~al.(2023)Corbett-Davies, Gaebler, Nilforoshan, Shroff, and Goel}]{corbett-davies2023measure}
Corbett-Davies, S.; Gaebler, J.~D.; Nilforoshan, H.; Shroff, R.; and Goel, S. 2023.
\newblock The Measure and Mismeasure of Fairness.
\newblock \emph{Journal of Machine Learning Research}, 24(312): 1--117.

\bibitem[{Delgado et~al.(2022)Delgado, de~Manuel, Parra, Moyano, Rueda, Guersenzvaig, Ausín, Cruz, Casacuberta, and Puyol}]{delgado2022bias}
Delgado, J.; de~Manuel, A.; Parra, I.; Moyano, C.; Rueda, J.; Guersenzvaig, A.; Ausín, T.; Cruz, M.; Casacuberta, D.; and Puyol, A. 2022.
\newblock Bias in Algorithms of AI Systems Developed for COVID-19: A Scoping Review.
\newblock \emph{Journal of Bioethical Inquiry}, 19(3): 407--419.

\bibitem[{Dwork et~al.(2012)Dwork, Hardt, Pitassi, Reingold, and Zemel}]{dwork2012fairness}
Dwork, C.; Hardt, M.; Pitassi, T.; Reingold, O.; and Zemel, R. 2012.
\newblock Fairness Through Awareness.
\newblock In \emph{Proceedings of the 3rd Innovations in Theoretical Computer Science Conference}, 214--226. New York, NY, USA: Association for Computing Machinery.

\bibitem[{Essink-Bot et~al.(2013)Essink-Bot, Lamkaddem, Jellema, Nielsen, and Stronks}]{essink2013interpreting}
Essink-Bot, M.-L.; Lamkaddem, M.; Jellema, P.; Nielsen, S.~S.; and Stronks, K. 2013.
\newblock Interpreting Ethnic Inequalities in Healthcare Consumption: A Conceptual Framework for Research.
\newblock \emph{The European Journal of Public Health}, 23(6): 922--926.

\bibitem[{Eubanks(2018)}]{eubanks2018automating}
Eubanks, V. 2018.
\newblock \emph{Automating Inequality: How High-Tech Tools Profile, Police, and Punish the Poor}.
\newblock New York: St. Martin's Press.
\newblock ISBN 9781250074317.

\bibitem[{{European Parliament and Council of the European Union}(2024)}]{euai2024}
{European Parliament and Council of the European Union}. 2024.
\newblock Regulation (EU) 2024/1689 of the European Parliament and of the Council of 13 June 2024 laying down harmonised rules on artificial intelligence and amending certain Union legislative acts (AI Act).
\newblock Official Journal of the European Union, L 2024/1689, 12 July 2024.

\bibitem[{Feldman et~al.(2015)Feldman, Friedler, Moeller, Scheidegger, and Venkatasubramanian}]{feldman2015certifying}
Feldman, M.; Friedler, S.~A.; Moeller, J.; Scheidegger, C.; and Venkatasubramanian, S. 2015.
\newblock Certifying and Removing Disparate Impact.
\newblock In \emph{Proceedings of the 21st ACM SIGKDD International Conference on Knowledge Discovery and Data Mining}, KDD '15, 259--268. New York, NY, USA: Association for Computing Machinery.
\newblock ISBN 978-1-4503-3664-2.

\bibitem[{Ferrara(2024)}]{ferrara2024genai}
Ferrara, E. 2024.
\newblock GenAI Against Humanity: Nefarious Applications of Generative Artificial Intelligence and Large Language Models.
\newblock \emph{Journal of Computational Social Science}, 7(1): 549--569.

\bibitem[{Fletcher, Nakeshimana, and Olubeko(2021)}]{fletcher2021addressing}
Fletcher, R.~R.; Nakeshimana, A.; and Olubeko, O. 2021.
\newblock Addressing Fairness, Bias, and Appropriate Use of Artificial Intelligence and Machine Learning in Global Health.
\newblock \emph{Frontiers in Artificial Intelligence}, 3: 561802.

\bibitem[{Flores, Kim, and Young(2024)}]{flores2024addressing}
Flores, L.; Kim, S.; and Young, S.~D. 2024.
\newblock Addressing Bias in Artificial Intelligence for Public Health Surveillance.
\newblock \emph{Journal of Medical Ethics}, 50(3): 190--194.

\bibitem[{Gagnier et~al.(2013)Gagnier, Kienle, Altman, Moher, Sox, and Riley}]{gagnier2013care}
Gagnier, J.~J.; Kienle, G.; Altman, D.~G.; Moher, D.; Sox, H.; and Riley, D. 2013.
\newblock The CARE Guidelines: Consensus-Based Clinical Case Reporting Guideline Development.
\newblock \emph{Global Advances in Health and Medicine}, 2(5): 38--43.

\bibitem[{Galanty et~al.(2024)Galanty, Luitse, Noteboom, Croon, Vlaar, Poell, Sanchez, Blanke, and Išgum}]{galanty2024assessing}
Galanty, M.; Luitse, D.; Noteboom, S.~H.; Croon, P.; Vlaar, A.~P.; Poell, T.; Sanchez, C.~I.; Blanke, T.; and Išgum, I. 2024.
\newblock Assessing the Documentation of Publicly Available Medical Image and Signal Datasets and Their Impact on Bias Using the BEAMRAD Tool.
\newblock \emph{Scientific Reports}, 14(1): 31846.

\bibitem[{Galhotra, Brun, and Meliou(2017)}]{galhotra2017fairness}
Galhotra, S.; Brun, Y.; and Meliou, A. 2017.
\newblock Fairness Testing: Testing Software for Discrimination.
\newblock In \emph{Proceedings of the 2017 11th Joint Meeting on Foundations of Software Engineering}, 498--510. ACM.

\bibitem[{Gebru et~al.(2021)Gebru, Morgenstern, Vecchione, Vaughan, Wallach, Daumé~III, and Crawford}]{gebru2021datasheets}
Gebru, T.; Morgenstern, J.; Vecchione, B.; Vaughan, J.~W.; Wallach, H.; Daumé~III, H.; and Crawford, K. 2021.
\newblock Datasheets for Datasets.
\newblock \emph{Communications of the ACM}, 64(12): 86--92.

\bibitem[{Gianfrancesco et~al.(2018)Gianfrancesco, Tamang, Yazdany, and Schmajuk}]{gianfrancesco2018potential}
Gianfrancesco, M.~A.; Tamang, S.; Yazdany, J.; and Schmajuk, G. 2018.
\newblock Potential Biases in Machine Learning Algorithms Using Electronic Health Record Data.
\newblock \emph{JAMA Internal Medicine}, 178(11): 1544--1547.

\bibitem[{Ginsberg et~al.(2009)Ginsberg, Mohebbi, Patel, Brammer, Smolinski, and Brilliant}]{Ginsberg_2009}
Ginsberg, J.; Mohebbi, M.~H.; Patel, R.~S.; Brammer, L.; Smolinski, M.~S.; and Brilliant, L. 2009.
\newblock Detecting influenza epidemics using search engine query data.
\newblock \emph{Nature}, 457(7232): 1012--1014.

\bibitem[{{Government of the Netherlands}(2024)}]{algoritmeregister2024}
{Government of the Netherlands}. 2024.
\newblock Algoritmeregister: Overview of algorithmic systems used by public institutions.
\newblock Available online, accessed 2025-03-22.

\bibitem[{Hardt, Price, and Srebro(2016)}]{hardt2016equality}
Hardt, M.; Price, E.; and Srebro, N. 2016.
\newblock Equality of Opportunity in Supervised Learning.
\newblock In \emph{Advances in Neural Information Processing Systems}, volume~29.

\bibitem[{Higgins et~al.(2011)Higgins, Altman, Gøtzsche, Jüni, Moher, Oxman, Savović, Schulz, Weeks, and Sterne}]{higgins2011cochrane}
Higgins, J. P.~T.; Altman, D.~G.; Gøtzsche, P.~C.; Jüni, P.; Moher, D.; Oxman, A.~D.; Savović, J.; Schulz, K.~F.; Weeks, L.; and Sterne, J. A.~C. 2011.
\newblock The Cochrane Collaboration’s Tool for Assessing Risk of Bias in Randomised Trials.
\newblock \emph{BMJ}, 343: d5928.

\bibitem[{Hillebrand et~al.(2020)Hillebrand, Khan, Peleja, and Oliver}]{hillebrand2020mobisenseus}
Hillebrand, M.; Khan, I.; Peleja, F.; and Oliver, N. 2020.
\newblock MobiSenseUs: Inferring Aggregate Objective and Subjective Well-being from Mobile Data.
\newblock In \emph{ECAI 2020: 24th European Conference on Artificial Intelligence}, 1818--1825. IOS Press.

\bibitem[{Holstege et~al.(2025)Holstege, van~de Geer, Leenes, and van Sluijs}]{holstege2025auditing}
Holstege, L.; van~de Geer, S.; Leenes, R.; and van Sluijs, J. 2025.
\newblock Auditing a Dutch Public Sector Risk Profiling Algorithm Using an Unsupervised Bias Detection Tool.
\newblock Preprint. Available at \url{https://arxiv.org/pdf/2502.01713}, arXiv:2502.01713v2.

\bibitem[{Holstein et~al.(2019)Holstein, Wortman~Vaughan, Daum{\'e}~III, Dud{\'\i}k, and Wallach}]{holstein2019improving}
Holstein, K.; Wortman~Vaughan, J.; Daum{\'e}~III, H.; Dud{\'\i}k, M.; and Wallach, H. 2019.
\newblock Improving Fairness in Machine Learning Systems: What Do Industry Practitioners Need?
\newblock In \emph{Proceedings of the 2019 CHI Conference on Human Factors in Computing Systems}, 1--16. ACM.

\bibitem[{Husereau et~al.(2013)Husereau, Drummond, Petrou, Carswell, Moher, Greenberg, Augustovski, Briggs, Mauskopf, and Loder}]{husereau2013cheers}
Husereau, D.; Drummond, M.; Petrou, S.; Carswell, C.; Moher, D.; Greenberg, D.; Augustovski, F.; Briggs, A.~H.; Mauskopf, J.; and Loder, E. 2013.
\newblock Consolidated Health Economic Evaluation Reporting Standards (CHEERS) Statement.
\newblock \emph{International Journal of Technology Assessment in Health Care}, 29(2): 117--122.

\bibitem[{Ibrahim et~al.(2021)Ibrahim, Liu, Denniston, Fraser, Keane, Faes, Geerts, Chambers, Corral, Lee, Wagner, and et~al.}]{ibrahim2021stardai}
Ibrahim, H.; Liu, X.; Denniston, A.~K.; Fraser, H.; Keane, P.~A.; Faes, L.; Geerts, B.; Chambers, D.; Corral, J.; Lee, A.~M.; Wagner, M.; and et~al. 2021.
\newblock STARD-AI: A Reporting Guideline for Studies Using Artificial Intelligence in Diagnostic Test Accuracy Studies.
\newblock \emph{BMJ Open}, 11(6): e041411.

\bibitem[{{IEEE}(2025)}]{ieee2025bias}
{IEEE}. 2025.
\newblock IEEE Standard for Algorithmic Bias Considerations.
\newblock IEEE Std 7003-2024.
\newblock Pages 1--59.

\bibitem[{Ikram et~al.(2014)Ikram, Kunst, Lamkaddem, and Stronks}]{ikram2014disease}
Ikram, U.~Z.; Kunst, A.~E.; Lamkaddem, M.; and Stronks, K. 2014.
\newblock The disease burden across different ethnic groups in Amsterdam, the Netherlands, 2011--2030.
\newblock \emph{The European Journal of Public Health}, 24(4): 600--605.

\bibitem[{Ilozumba et~al.(2022)Ilozumba, Koster, Syurina, and Ebuenyi}]{ilozumba2022ethnic}
Ilozumba, O.; Koster, T.~S.; Syurina, E.~V.; and Ebuenyi, I. 2022.
\newblock Ethnic minority experiences of mental health services in the Netherlands: An exploratory study.
\newblock \emph{BMC Research Notes}, 15(1): 266.

\bibitem[{Jiang et~al.(2017)Jiang, Jiang, Zhi, Dong, Li, Ma, Wang, Dong, Shen, and Wang}]{jiang2017artificial}
Jiang, F.; Jiang, Y.; Zhi, H.; Dong, Y.; Li, H.; Ma, S.; Wang, Y.; Dong, Q.; Shen, H.; and Wang, Y. 2017.
\newblock Artificial intelligence in healthcare: past, present and future.
\newblock \emph{Stroke and Vascular Neurology}, 2(4): 230--243.

\bibitem[{Kilkenny et~al.(2010)Kilkenny, Browne, Cuthill, Emerson, and Altman}]{kilkenny2010arrive}
Kilkenny, C.; Browne, W.~J.; Cuthill, I.~C.; Emerson, M.; and Altman, D.~G. 2010.
\newblock Improving bioscience research reporting: the ARRIVE guidelines for reporting animal research.
\newblock \emph{PLoS Biology}, 8(6): e1000412.

\bibitem[{Kroneman et~al.(2016)Kroneman, Boerma, van~den Berg, Groenewegen, de~Jong, and van Ginneken}]{kroneman2016netherlands}
Kroneman, M.; Boerma, W.; van~den Berg, M.; Groenewegen, P.; de~Jong, J.; and van Ginneken, E. 2016.
\newblock Netherlands: health system review.
\newblock \emph{Health Systems in Transition}, 18(2): 1--240.

\bibitem[{Kusner et~al.(2017)Kusner, Loftus, Russell, and Silva}]{kusner2017counterfactual}
Kusner, M.~J.; Loftus, J.; Russell, C.; and Silva, R. 2017.
\newblock Counterfactual Fairness.
\newblock In \emph{Advances in Neural Information Processing Systems}, volume~30.

\bibitem[{Lekadir et~al.(2025)Lekadir, Frangi, Porras, Glocker, Cintas, Langlotz, Weicken, Asselbergs, Prior, Collins, Kaissis, Tsakou, Buvat, Kalpathy-Cramer, Mongan, Schnabel, Kushibar, Riklund, Marias, Amugongo, Fromont, Maier-Hein, Cerdá-Alberich, Martí-Bonmatí, Cardoso, Bobowicz, Shabani, Tsiknakis, Zuluaga, Fritzsche, Camacho, Linguraru, Wenzel, De~Bruijne, Tolsgaard, Goisauf, Cano~Abadía, Papanikolaou, Lazrak, Pujol, Osuala, Napel, Colantonio, Joshi, Klein, Aussó, Rogers, Salahuddin, Starmans, and Consortium}]{lekadir2025futureai}
Lekadir, K.; Frangi, A.~F.; Porras, A.~R.; Glocker, B.; Cintas, C.; Langlotz, C.~P.; Weicken, E.; Asselbergs, F.~W.; Prior, F.; Collins, G.~S.; Kaissis, G.; Tsakou, G.; Buvat, I.; Kalpathy-Cramer, J.; Mongan, J.; Schnabel, J.~A.; Kushibar, K.; Riklund, K.; Marias, K.; Amugongo, L.~M.; Fromont, L.~A.; Maier-Hein, L.; Cerdá-Alberich, L.; Martí-Bonmatí, L.; Cardoso, M.~J.; Bobowicz, M.; Shabani, M.; Tsiknakis, M.; Zuluaga, M.~A.; Fritzsche, M.-C.; Camacho, M.; Linguraru, M.~G.; Wenzel, M.; De~Bruijne, M.; Tolsgaard, M.~G.; Goisauf, M.; Cano~Abadía, M.; Papanikolaou, N.; Lazrak, N.; Pujol, O.; Osuala, R.; Napel, S.; Colantonio, S.; Joshi, S.; Klein, S.; Aussó, S.; Rogers, W.~A.; Salahuddin, Z.; Starmans, M. P.~A.; and Consortium, F.-A. 2025.
\newblock FUTURE-AI: International consensus guideline for trustworthy and deployable artificial intelligence in healthcare.
\newblock \emph{BMJ (Clinical research ed.)}, 388: e081554.

\bibitem[{Liu et~al.(2020)Liu, Cruz~Rivera, Moher, Calvert, and Denniston}]{liu2020consortai}
Liu, X.; Cruz~Rivera, S.; Moher, D.; Calvert, M.~J.; and Denniston, A.~K. 2020.
\newblock Reporting guidelines for clinical trial reports for interventions involving artificial intelligence: the CONSORT-AI extension.
\newblock \emph{Nature Medicine}, 26: 1364--1374.

\bibitem[{Madaio et~al.(2020)Madaio, Stark, Wortman~Vaughan, and Wallach}]{madaio2020co}
Madaio, M.~A.; Stark, L.; Wortman~Vaughan, J.; and Wallach, H. 2020.
\newblock Co-Designing Checklists to Understand Organizational Challenges and Opportunities around Fairness in AI.
\newblock In \emph{Proceedings of the 2020 CHI Conference on Human Factors in Computing Systems}, CHI '20, 1--14. New York, NY, USA: Association for Computing Machinery.

\bibitem[{Margetts and Naumann(2017)}]{margetts2017government}
Margetts, H.; and Naumann, A. 2017.
\newblock Government as a Platform: What Can Estonia Show the World?
\newblock Research paper, University of Oxford.

\bibitem[{Mehrabi et~al.(2021)Mehrabi, Morstatter, Saxena, Lerman, and Galstyan}]{mehrabi2021survey}
Mehrabi, N.; Morstatter, F.; Saxena, N.; Lerman, K.; and Galstyan, A. 2021.
\newblock A Survey on Bias and Fairness in Machine Learning.
\newblock \emph{ACM Computing Surveys}, 54(6): 1--35.

\bibitem[{Mhasawade, Zhao, and Chunara(2021)}]{mhasawade2021machine}
Mhasawade, V.; Zhao, Y.; and Chunara, R. 2021.
\newblock Machine Learning and Algorithmic Fairness in Public and Population Health.
\newblock \emph{Nature Machine Intelligence}, 3(8): 659--666.

\bibitem[{{Ministry of Infrastructure and Water Management}(2024)}]{miniwm2024aiimpact}
{Ministry of Infrastructure and Water Management}. 2024.
\newblock AI Impact Assessment: The Tool for a Responsible AI Project.
\newblock Technical report, Government of the Netherlands.

\bibitem[{{Ministry of the Interior and Kingdom Relations}(2024)}]{minbzk2024genai}
{Ministry of the Interior and Kingdom Relations}. 2024.
\newblock Government-wide Vision on Generative AI of the Netherlands.
\newblock Technical report, Government of the Netherlands.

\bibitem[{Mitchell et~al.(2019)Mitchell, Wu, Zaldivar, Barnes, Vasserman, Hutchinson, Spitzer, Raji, and Gebru}]{mitchell2019model}
Mitchell, M.; Wu, S.; Zaldivar, A.; Barnes, P.; Vasserman, L.; Hutchinson, B.; Spitzer, E.; Raji, I.~D.; and Gebru, T. 2019.
\newblock Model Cards for Model Reporting.
\newblock In \emph{Proceedings of the 2019 Conference on Fairness, Accountability, and Transparency (FAT* 2019)}, 220--229. New York, NY, USA: Association for Computing Machinery.

\bibitem[{Mittelstadt et~al.(2016)Mittelstadt, Allo, Taddeo, Wachter, and Floridi}]{mittelstadt2016ethics}
Mittelstadt, B.~D.; Allo, P.; Taddeo, M.; Wachter, S.; and Floridi, L. 2016.
\newblock The Ethics of Algorithms: Mapping the Debate.
\newblock \emph{Big Data \& Society}, 3(2): 2053951716679679.

\bibitem[{Mongan, Moy, and Kahn(2020)}]{mongan2020claim}
Mongan, J.; Moy, L.; and Kahn, C. E.~J. 2020.
\newblock Checklist for Artificial Intelligence in Medical Imaging (CLAIM): A Guide for Authors and Reviewers.
\newblock \emph{Radiology: Artificial Intelligence}, 2(2): e200029.

\bibitem[{Moons et~al.(2025)Moons, Damen, Kaul, Riley, Reitsma, van Smeden, Wolff et~al.}]{moons2025probastai}
Moons, K. G.~M.; Damen, J. A.~A.; Kaul, T.; Riley, R.~D.; Reitsma, J.~B.; van Smeden, M.; Wolff, R.~F.; et~al. 2025.
\newblock {PROBAST}+{AI}: an updated quality, risk of bias, and applicability assessment tool for prediction models using regression or artificial intelligence methods.
\newblock \emph{BMJ}, 388: e082505.

\bibitem[{Moons et~al.(2019)Moons, Wolff, Riley, Whiting, Westwood, Collins, Reitsma, Kleijnen, and Mallett}]{moons2019probast}
Moons, K. G.~M.; Wolff, R.~F.; Riley, R.~D.; Whiting, P.~F.; Westwood, M.; Collins, G.~S.; Reitsma, J.~B.; Kleijnen, J.; and Mallett, S. 2019.
\newblock PROBAST: A Tool to Assess Risk of Bias and Applicability of Prediction Model Studies: Explanation and Elaboration.
\newblock \emph{Annals of Internal Medicine}, 170(1): W1--W33.

\bibitem[{Morgenstern et~al.(2020)Morgenstern, Buajitti, O’Neill, Piggott, Goel, Fridman, Kornas, and Rosella}]{morgenstern2020predicting}
Morgenstern, J.~D.; Buajitti, E.; O’Neill, M.; Piggott, T.; Goel, V.; Fridman, D.; Kornas, K.; and Rosella, L.~C. 2020.
\newblock Predicting Population Health with Machine Learning: A Scoping Review.
\newblock \emph{BMJ Open}, 10(10): e037860.

\bibitem[{Obermeyer et~al.(2019)Obermeyer, Powers, Vogeli, and Mullainathan}]{obermeyer2019dissecting}
Obermeyer, Z.; Powers, B.; Vogeli, C.; and Mullainathan, S. 2019.
\newblock Dissecting Racial Bias in an Algorithm Used to Manage the Health of Populations.
\newblock \emph{Science}, 366(6464): 447--453.

\bibitem[{O'Brien et~al.(2014)O'Brien, Harris, Beckman, Reed, and Cook}]{obrien2014standards}
O'Brien, B.~C.; Harris, I.~B.; Beckman, T.~J.; Reed, D.~A.; and Cook, D.~A. 2014.
\newblock Standards for Reporting Qualitative Research: A Synthesis of Recommendations.
\newblock \emph{Academic Medicine}, 89(9): 1245--1251.

\bibitem[{Page et~al.(2021)Page, McKenzie, Bossuyt, Boutron, Hoffmann, Mulrow, Shamseer, Tetzlaff, Akl, Brennan, Chou, Glanville, Grimshaw, Hróbjartsson, Lalu, Li, Loder, Mayo-Wilson, McDonald, McGuinness, Stewart, Thomas, Tricco, Welch, Whiting, and Moher}]{page2021prisma}
Page, M.~J.; McKenzie, J.~E.; Bossuyt, P.~M.; Boutron, I.; Hoffmann, T.~C.; Mulrow, C.~D.; Shamseer, L.; Tetzlaff, J.~M.; Akl, E.~A.; Brennan, S.~E.; Chou, R.; Glanville, J.; Grimshaw, J.~M.; Hróbjartsson, A.; Lalu, M.~M.; Li, T.; Loder, E.~W.; Mayo-Wilson, E.; McDonald, S.; McGuinness, L.~A.; Stewart, L.~A.; Thomas, J.; Tricco, A.~C.; Welch, V.~A.; Whiting, P.; and Moher, D. 2021.
\newblock The PRISMA 2020 Statement: An Updated Guideline for Reporting Systematic Reviews.
\newblock \emph{BMJ}, 372: n71.

\bibitem[{Patel et~al.(2015)Patel, Shah, Thakkar, and Kotecha}]{patel2015predicting}
Patel, J.; Shah, S.; Thakkar, P.; and Kotecha, K. 2015.
\newblock Predicting Stock Market Index Using Fusion of Machine Learning Techniques.
\newblock \emph{Expert Systems with Applications}, 42(4): 2162--2172.

\bibitem[{Rajkomar et~al.(2018)Rajkomar, Hardt, Howell, Corrado, and Chin}]{rajkomar2018ensuring}
Rajkomar, A.; Hardt, M.; Howell, M.~D.; Corrado, G.; and Chin, M.~H. 2018.
\newblock Ensuring Fairness in Machine Learning to Advance Health Equity.
\newblock \emph{Annals of Internal Medicine}, 169(12): 866--872.

\bibitem[{Raza(2023)}]{raza2023connecting}
Raza, S. 2023.
\newblock Connecting Fairness in Machine Learning with Public Health Equity.
\newblock In \emph{Proceedings of the 2023 IEEE 11th International Conference on Healthcare Informatics (ICHI)}, 704--708. IEEE.

\bibitem[{Raza et~al.(2024)Raza, Shaban-Nejad, Dolatabadi, and Mamiya}]{raza2024exploring}
Raza, S.; Shaban-Nejad, A.; Dolatabadi, E.; and Mamiya, H. 2024.
\newblock Exploring Bias and Prediction Metrics to Characterise the Fairness of Machine Learning for Equity-Centered Public Health Decision-Making: A Narrative Review.
\newblock \emph{IEEE Access}, 12: 180815--180829.

\bibitem[{Rostamzadeh et~al.(2022)Rostamzadeh, Mincu, Roy, Smart, Wilcox, Pushkarna, Schrouff, Amironesei, Moorosi, and Heller}]{rostamzadeh2022healthsheet}
Rostamzadeh, N.; Mincu, D.; Roy, S.; Smart, A.; Wilcox, L.; Pushkarna, M.; Schrouff, J.; Amironesei, R.; Moorosi, N.; and Heller, K. 2022.
\newblock Healthsheet: Development of a Transparency Artifact for Health Datasets.
\newblock In \emph{Proceedings of the 2022 ACM Conference on Fairness, Accountability, and Transparency}, 1943--1961. Association for Computing Machinery.

\bibitem[{Saleiro et~al.(2018)Saleiro, Kuester, Hinkson, London, Stevens, Anisfeld, Rodolfa, and Ghani}]{saleiro2018aequitas}
Saleiro, P.; Kuester, B.; Hinkson, L.; London, J.; Stevens, A.; Anisfeld, A.; Rodolfa, K.~T.; and Ghani, R. 2018.
\newblock Aequitas: A Bias and Fairness Audit Toolkit.
\newblock arXiv preprint arXiv:1811.05577.

\bibitem[{Selbst et~al.(2019)Selbst, boyd, Friedler, Venkatasubramanian, and Vertesi}]{selbst2019fairness}
Selbst, A.~D.; boyd, d.; Friedler, S.~A.; Venkatasubramanian, S.; and Vertesi, J. 2019.
\newblock Fairness and Abstraction in Sociotechnical Systems.
\newblock In \emph{Proceedings of the 2019 Conference on Fairness, Accountability, and Transparency (FAT*)}, 59--68. Association for Computing Machinery.

\bibitem[{Sikstrom et~al.(2022)Sikstrom, Maslej, Hui, Findlay, Buchman, and Hill}]{sikstrom2022conceptualising}
Sikstrom, L.; Maslej, M.~M.; Hui, K.; Findlay, Z.; Buchman, D.~Z.; and Hill, S.~L. 2022.
\newblock Conceptualising Fairness: Three Pillars for Medical Algorithms and Health Equity.
\newblock \emph{BMJ Health \& Care Informatics}, 29(1): e100459.

\bibitem[{Suresh and Guttag(2019)}]{suresh2019framework}
Suresh, H.; and Guttag, J.~V. 2019.
\newblock A Framework for Understanding Unintended Consequences of Machine Learning.
\newblock arXiv preprint arXiv:1901.10002.

\bibitem[{Teunissen et~al.(2015)Teunissen, Van~Bavel, Van~den Driessen~Mareeuw, MacFarlane, Van Weel-Baumgarten, Van~den Muijsenbergh, and Van~Weel}]{teunissen2015mental}
Teunissen, E.; Van~Bavel, E.; Van~den Driessen~Mareeuw, F.; MacFarlane, A.; Van Weel-Baumgarten, E.; Van~den Muijsenbergh, M.; and Van~Weel, C. 2015.
\newblock Mental Health Problems of Undocumented Migrants in the Netherlands: A Qualitative Exploration of Recognition, Recording, and Treatment by General Practitioners.
\newblock \emph{Scandinavian Journal of Primary Health Care}, 33(2): 82--90.

\bibitem[{Thomasian, Eickhoff, and Adashi(2021)}]{thomasian2021advancing}
Thomasian, N.~M.; Eickhoff, C.; and Adashi, E.~Y. 2021.
\newblock Advancing Health Equity with Artificial Intelligence.
\newblock \emph{Journal of Public Health Policy}, 42(4): 602--611.

\bibitem[{Tong, Sainsbury, and Craig(2007)}]{tong2007coreq}
Tong, A.; Sainsbury, P.; and Craig, J. 2007.
\newblock Consolidated Criteria for Reporting Qualitative Research (COREQ): A 32-Item Checklist for Interviews and Focus Groups.
\newblock \emph{International Journal for Quality in Health Care}, 19(6): 349--357.

\bibitem[{Tram{\`e}r et~al.(2017)Tram{\`e}r, Atlidakis, Geambasu, Hsu, Hubaux, Humbert, Juels, and Lin}]{tramer2017fairtest}
Tram{\`e}r, F.; Atlidakis, V.; Geambasu, R.; Hsu, D.; Hubaux, J.-P.; Humbert, M.; Juels, A.; and Lin, H. 2017.
\newblock FairTest: Discovering Unwarranted Associations in Data-Driven Applications.
\newblock In \emph{2017 IEEE European Symposium on Security and Privacy (EuroS\&P)}, 401--416. IEEE.

\bibitem[{Tsai et~al.(2022)Tsai, Arik, Jacobson, Yoon, Yoder, Sava, Mitchell, Graham, and Pfister}]{tsai2022algorithmic}
Tsai, T.~C.; Arik, S.; Jacobson, B.~H.; Yoon, J.; Yoder, N.; Sava, D.; Mitchell, M.; Graham, G.; and Pfister, T. 2022.
\newblock Algorithmic Fairness in Pandemic Forecasting: Lessons from COVID-19.
\newblock \emph{NPJ Digital Medicine}, 5(1): 59.

\bibitem[{Verma and Rubin(2018)}]{verma2018fairness}
Verma, S.; and Rubin, J. 2018.
\newblock Fairness Definitions Explained.
\newblock In \emph{Proceedings of the International Workshop on Software Fairness}, 1--7. ACM.

\bibitem[{Vollmer et~al.(2020)Vollmer, Mateen, Bohner, Kir{\'a}ly, Ghani, Jonsson, Cumbers, Jonas, McAllister, Myles, and et~al.}]{vollmer2020machine}
Vollmer, S.; Mateen, B.~A.; Bohner, G.; Kir{\'a}ly, F.~J.; Ghani, R.; Jonsson, P.; Cumbers, S.; Jonas, A.; McAllister, K.~S.; Myles, P.; and et~al. 2020.
\newblock Machine Learning and Artificial Intelligence Research for Patient Benefit: 20 Critical Questions on Transparency, Replicability, Ethics, and Effectiveness.
\newblock \emph{BMJ}, 368: l6927.

\bibitem[{von Elm et~al.(2007)von Elm, Altman, Egger, Pocock, G{\o}tzsche, and Vandenbroucke}]{vonelm2007strobe}
von Elm, E.; Altman, D.~G.; Egger, M.; Pocock, S.~J.; G{\o}tzsche, P.~C.; and Vandenbroucke, J.~P. 2007.
\newblock The Strengthening the Reporting of Observational Studies in Epidemiology (STROBE) Statement: Guidelines for Reporting Observational Studies.
\newblock \emph{The Lancet}, 370(9596): 1453--1457.

\bibitem[{Wesson et~al.(2022)Wesson, Hswen, Valdes, Stojanovski, and Handley}]{wesson2022risks}
Wesson, P.; Hswen, Y.; Valdes, G.; Stojanovski, K.; and Handley, M.~A. 2022.
\newblock Risks and Opportunities to Ensure Equity in the Application of Big Data Research in Public Health.
\newblock \emph{Annual Review of Public Health}, 43(1): 59--78.

\bibitem[{Wexler et~al.(2020)Wexler, Pushkarna, Bolukbasi, Wattenberg, Viégas, and Wilson}]{wexler2019whatif}
Wexler, J.; Pushkarna, M.; Bolukbasi, T.; Wattenberg, M.; Viégas, F.; and Wilson, J. 2020.
\newblock The What-If Tool: Interactive Probing of Machine Learning Models.
\newblock \emph{IEEE Transactions on Visualization and Computer Graphics}, 26(1): 56--65.

\bibitem[{Wiemken and Kelley(2020)}]{wiemken2020machine}
Wiemken, T.~L.; and Kelley, R.~R. 2020.
\newblock Machine Learning in Epidemiology and Health Outcomes Research.
\newblock \emph{Annual Review of Public Health}, 41: 21--36.

\bibitem[{Wiśniewski and Biecek(2022)}]{wisniewski2022fairmodels}
Wiśniewski, J.; and Biecek, P. 2022.
\newblock fairmodels: A Flexible Tool for Bias Detection, Visualization, and Mitigation in Binary Classification Models.
\newblock \emph{The R Journal}, 14(1): 227--243.

\bibitem[{Wolff et~al.(2019)Wolff, Moons, Riley, Whiting, Westwood, Collins, Reitsma, Kleijnen, and Mallett}]{wolff2019probast}
Wolff, R.~F.; Moons, K. G.~M.; Riley, R.~D.; Whiting, P.~F.; Westwood, M.; Collins, G.~S.; Reitsma, J.~B.; Kleijnen, J.; and Mallett, S. 2019.
\newblock PROBAST: A Tool to Assess the Risk of Bias and Applicability of Prediction Model Studies.
\newblock \emph{Annals of Internal Medicine}, 170(1): 51--58.

\bibitem[{{World Health Organization}(2021)}]{who2021ethics}
{World Health Organization}. 2021.
\newblock Ethics and Governance of Artificial Intelligence for Health.
\newblock Accessed 2025-05-22.

\bibitem[{Xu et~al.(2022)Xu, Xiao, Wang, Ning, Shenkman, Bian, and Wang}]{xu2022algorithmic}
Xu, J.; Xiao, Y.; Wang, W.~H.; Ning, Y.; Shenkman, E.~A.; Bian, J.; and Wang, F. 2022.
\newblock Algorithmic Fairness in Computational Medicine.
\newblock \emph{eBioMedicine}, 84: 104250.

\end{thebibliography}

\appendix

\section{Appendix}

\renewcommand{\thetable}{\thesection.\arabic{table}}
\setcounter{table}{0}

\begin{table*}[h]
\centering
\begin{tabular}{p{0.3\linewidth}p{0.6\linewidth}}
\toprule
\textbf{Question} & \textbf{Full Question Text} \\
\midrule
Q1. Data Bias & To what extent in terms of depth is the potential risk of data bias discussed in the paper? Note: For example, issues related to data representativeness, collection processes, or sampling biases? \\
Q2. Model Bias & To what extent in terms of depth is the potential risk of algorithmic/model bias discussed in the paper? Note: For example, disparities in performance across demographic groups or systematic errors in predictions? \\
Q3. ML Fairness & Are the potential risks of bias discussed in the context of fairness in machine learning? \\
Q4. Societal Impact & To what extent does the paper discuss the potential societal impacts of the algorithm/model, particularly regarding the expected benefits and potential harms for different stakeholder groups? \\
\bottomrule
\end{tabular}
\caption{RABAT Category 1: Extent of Bias Discussion. Research question: How do the studies engage with bias in ML in terms of origins, fairness, and impact?}
\label{tab:rabat-cat1}
\end{table*}

\begin{table*}[h]
\centering
\begin{tabular}{p{0.3\linewidth}p{0.6\linewidth}}
\toprule
\textbf{Question} & \textbf{Full Question Text} \\
\midrule
Q5. Subgroups at Risk & Does the paper identify specific health characteristics or sociodemographic groups that might be at risk of experiencing disproportionate negative effects due to potential biases in the algorithm? \\
Q6. Sampling and Missing Data & Was the data/participant sampling method clearly described in the paper? Were exclusions and inclusions of data points explicit and justified? \\
Q7. Sensitive Attributes & Does the paper identify and describe the demographic characteristics (or ‘sensitive attributes’) of the stakeholders who might be negatively affected by the algorithm/model (e.g., race, gender, age, disability status)? \\
\bottomrule
\end{tabular}
\caption{RABAT Category 2: Identification of Subgroups at Risk. Research question: Does the paper identify specific subgroups at risk of disproportionate harms?}
\label{tab:rabat-cat2}
\end{table*}

\begin{table*}[h]
\centering
\begin{tabular}{p{0.3\linewidth}p{0.6\linewidth}}
\toprule
\textbf{Question} & \textbf{Full Question Text} \\
\midrule
Q8. Bias Articulation & To what extent in terms of length and structure does the paper articulate and describe the potential risks of data and/or model bias? \\
Q9. Harm Transparency & Is there transparent reporting on how fairness-related harms were identified, assessed, and mitigated throughout the algorithm/model lifecycle? \\
Q10. Informed Consent & Does the paper explain how meaningful informed consent was obtained during data collection? (Write NA if not applicable.) \\
\bottomrule
\end{tabular}
\caption{RABAT Category 3: Reporting Transparency. Research question: How do the selected studies report risks of bias?}
\label{tab:rabat-cat3}
\end{table*}

\begin{table*}[ht]
\centering
\begin{tabular}{@{}p{0.1\linewidth}p{0.2\linewidth}p{0.3\linewidth}p{0.3\linewidth}@{}}
\toprule
\textbf{Score} & \textbf{Criteria} & \textbf{Detailed Description} & \textbf{Example Scoring Notes} \\
\midrule
0-Absent      & No mention of data bias.                  & No discussion of data representativeness, collection biases, or dataset limitations.           & No mention of data representativeness or sampling bias in model development. \\
1-Minimal    & Brief mention, no depth (e.g., one or two sentences). & Acknowledges dataset limitations (e.g., missing data, small sample sizes) but does not analyze bias or its implications. & “This dataset may have limitations due to missing values.” (p.\ X) \\
2-Moderate   & Covers key aspects but lacks depth (e.g., a short paragraph).  & Identifies data bias risks (e.g., sampling bias, unbalanced datasets) but lacks detailed analysis or mitigation strategies. & “The dataset primarily includes data from urban hospitals, which may not generalize to rural populations.” (p.\ X) \\
3-Extensive  & Fully developed discussion (e.g., multiple paragraphs or a dedicated section). & Examines data bias origins, discusses subgroup disparities, and proposes mitigation strategies.     & “To address potential bias, we used stratified sampling to ensure demographic balance in the dataset.” (p.\ X) \\
\bottomrule
\end{tabular}
\caption{RABAT grading scale used to score extent of bias discussion, subgroup identification, and reporting transparency in PH+ML studies.}
\label{tab:rabat-grading}
\end{table*}

\begin{table*}[h]
\centering
\begin{tabular}{@{}p{0.05\linewidth}p{0.25\linewidth}p{0.60\linewidth}@{}}
\toprule
\textbf{Q\#} & \textbf{RABAT Question (Short Name)} & \textbf{Primary Source Inspiration(s)} \\
\midrule
Q1 & Data Bias & Cochrane reporting bias; PROBAST on predictor definition and measurement; MS RAI on dataset limitations and representativeness \\
Q2 & Model Bias & PROBAST on model evaluation and discrimination; MS RAI on error analysis and performance disparities \\
Q3 & ML Fairness & MS RAI fairness framing, harm types, and stakeholder impacts \\
Q4 & Societal Impact & MS RAI system’s role in society, power asymmetries, and trade-offs between harms and benefits \\
Q5 & Subgroups at Risk & PROBAST participant sampling and subgroup clarity; MS RAI expected stakeholders and fairness considerations \\
Q6 & Sampling and Missing Data & Cochrane selection bias; PROBAST on inclusion/exclusion criteria and missing data; MS RAI data completeness \\
Q7 & Sensitive Attributes & PROBAST on predictor relevance; MS RAI on demographic characteristics and protected attributes \\
Q8 & Bias Articulation & Cochrane on reporting clarity; PROBAST on overall bias synthesis; MS RAI on structured bias reflection \\
Q9 & Harm Transparency & PROBAST on analytical decisions and bias control; MS RAI on fairness interventions and mitigation strategies \\
Q10 & Informed Consent & MS RAI on consent processes, data subject awareness, and opt-out provisions \\
\bottomrule
\end{tabular}
\caption{Mapping of RABAT questions from conceptual foundations in Cochrane Risk of Bias, PROBAST (2019), and the Microsoft Responsible AI checklist (MS RAI).}
\label{tab:rabat-mapping}
\end{table*}

\begin{table*}[h]
\centering
\begin{tabular}{@{}p{0.35\linewidth}p{0.40\linewidth}p{0.15\linewidth}@{}}
\toprule
\textbf{Design Thinking Stage (Short Definition)} & \textbf{ACAR Stage (Definition)} & \textbf{RABAT Qs} \\
\midrule
\textbf{Empathize}: Understand user needs through observation and engagement. & \textbf{Awareness}: Recognize that AB may arise from data, models, or context. Consider fairness relevance, affected groups, and societal impact. & Q1, Q2, Q4 \\
\addlinespace
\textbf{Define}: Frame the problem clearly to guide targeted solutions. & \textbf{Conceptualization}: Define fairness and subgroup risk. Link to study design and PH context. & Q3, Q5 \\
\addlinespace
\textbf{Ideate}: Generate a range of solution strategies. & \textbf{Application}: Apply fairness-aware methods in data and modeling workflows. & Q6, Q7, Q9 \\
\addlinespace
\textbf{Prototype}: Build simplified versions for testing and refinement. & \textbf{Application (continued)}: Includes experimental testing of subgroup analysis or mitigation methods. & Q6, Q7, Q9 \\
\addlinespace
\textbf{Test}: Evaluate whether the solution addresses user and contextual needs. & \textbf{Reporting}: Report AB risks, subgroup findings, and fairness-related limitations. & Q8, Q9, Q10 \\
\bottomrule
\end{tabular}
\caption{Mapping Design Thinking Stages to ACAR Framework and RABAT Questions}
\label{tab:dt_acar_mapping}
\end{table*}

\begin{table*}[h]
\centering
\begin{tabular}{@{}p{0.01\linewidth}p{0.08\linewidth}p{0.02\linewidth}p{0.82\linewidth}@{}}
\toprule
\textbf{\#} & \textbf{Author} & \textbf{Year} & \textbf{Title} \\
\midrule
1  & van Vuuren      & 2021 & Comparing machine learning to a rule-based approach for predicting suicidal behavior among adolescents: Results from a longitudinal population-based survey \\
2  & Schinkel        & 2022 & Diagnostic stewardship for blood cultures in the emergency department: A multicenter validation and prospective evaluation of a machine learning prediction tool \\
3  & Viljanen        & 2022 & A machine learning approach to small area estimation: predicting the health, housing and well-being of the population of Netherlands \\
4  & Fussenich       & 2021 & Mapping chronic disease prevalence based on medication use and socio-demographic variables: an application of LASSO on administrative data sources in healthcare in the Netherlands \\
5  & Velders         & 2021 & Improvements in air quality in the Netherlands during the corona lockdown based on observations and model simulations \\
6  & Liu             & 2022 & Automated food safety early warning system in the dairy supply chain using machine learning \\
7  & Wang            & 2022 & Designing a monitoring program for aflatoxin B1 in feed products using machine learning \\
8  & Zehnder         & 2023 & Machine Learning for Detecting Virus Infection Hotspots Via Wastewater-Based Epidemiology: The Case of SARS-CoV-2 RNA \\
9  & Hoekstra        & 2023 & Predicting self-perceived general health status using machine learning: an external exposome study \\
10 & Meppelink       & 2021 & Reliable or not? An automated classification of webpages about early childhood vaccination using supervised machine learning \\
11 & De Nijs         & 2021 & Individualized prediction of three- and six-year outcomes of psychosis in a longitudinal multicenter study: a machine learning approach \\
12 & Spijker         & 2023 & A machine learning based modelling framework to predict nitrate leaching from agricultural soils across the Netherlands \\
13 & Loef            & 2022 & Using random forest to identify longitudinal predictors of health in a 30-year cohort study \\
14 & Gogishvili      & 2023 & Discovery of novel CSF biomarkers to predict progression in dementia using machine learning \\
15 & Achterberg      & 2022 & Comparing the accuracy of several network-based COVID-19 prediction algorithms \\
16 & Qian            & 2021 & Integrating Expert ODEs into Neural ODEs: Pharmacology and Disease Progression \\
17 & Ramos           & 2021 & Predicting Success of a Digital Self-Help Intervention for Alcohol and Substance Use With Machine Learning \\
18 & Kers            & 2022 & Deep learning-based classification of kidney transplant pathology: a retrospective, multicentre, proof-of-concept study \\
19 & Krusemann       & 2021 & Comprehensive overview of common e-liquid ingredients and how they can be used to predict an e-liquid’s flavour category \\
20 & Ouwerkerk       & 2023 & Multiomics Analysis Provides Novel Pathways Related to Progression of Heart Failure \\
21 & Sustersic       & 2021 & Epidemiological Predictive Modeling of COVID-19 Infection: Development, Testing, and Implementation on the Population of the Benelux Union \\
22 & Molenaar        & 2024 & Predicting population-level vulnerability among pregnant women using routinely collected data and the added relevance of self-reported data \\
23 & Pelt            & 2024 & Building machine learning prediction models for well-being using predictors from the exposome and genome in a population cohort \\
24 & Schut           & 2024 & Development and evaluation of regression tree models for predicting in-hospital mortality of a national registry of COVID-19 patients over six pandemic surges \\
25 & Kurucz          & 2024 & Prediction of emergency department presentations for acute coronary syndrome using a machine learning approach \\
26 & Hernandez       & 2024 & Using XGBoost and SHAP to explain citizens’ differences in policy support for reimposing COVID-19 measures in the Netherlands \\
27 & Rydin           & 2025 & Predicting incident cardio-metabolic disease among persons with and without depressive and anxiety disorders: a machine learning approach \\
28 & Sonnaville      & 2024 & Predicting long-term neurocognitive outcome after pediatric intensive care unit admission for bronchiolitis—preliminary exploration of the potential of machine learning \\
29 & Helms           & 2024 & Determinants of Citizens’ Intention to Participate in Self-Led Contact Tracing: Cross-Sectional Online Questionnaire Study \\
30 & Mughini-Gras    & 2025 & Source attribution of Listeria monocytogenes in the Netherlands \\
31 & Warmbrunn       & 2024 & Networks of gut bacteria relate to cardiovascular disease in a multi-ethnic population: the HELIUS study \\
32 & Westerbeek      & 2024 & SeNiors empOWered via Big Data to Joint-Manage Their Medication-Related Risk of Falling in Primary Care: The SNOWDROP Project \\
33 & Smit            & 2024 & Past or Present; Which Exposures Predict Metabolomic Aging Better? The Doetinchem Cohort Study \\
34 & Pellemans       & 2024 & Automated Behavioral Coding to Enhance the Effectiveness of Motivational Interviewing in a Chat-Based Suicide Prevention Helpline: Secondary Analysis of a Clinical Trial \\
35 & Kanning         & 2024 & Prediction of aneurysmal subarachnoid hemorrhage in comparison with other stroke types using routine care data \\
\bottomrule
\end{tabular}
\caption{Included studies in systematic literature review.}
\label{tab:included-studies}
\end{table*}

\end{document}